%% file: main.tex
\documentclass[11pt]{article}

\usepackage[margin=1in]{geometry}
\usepackage{times}
\usepackage{microtype}
\usepackage{amsmath,amssymb}
\usepackage{booktabs}
\usepackage{graphicx}
\usepackage{float}
\usepackage{subcaption}
\usepackage{placeins}
\usepackage{array}
\usepackage{tabularx}
\usepackage{xcolor}
\usepackage[hidelinks]{hyperref}
\usepackage[numbers,sort&compress]{natbib}
\usepackage{enumitem}

\graphicspath{
  {figures/}
  {../../phase2-experiment/results/analysis_outputs/gkd_dynamics/figures/}
  {../../phase2-experiment/results/analysis_outputs/e1_boundary/}
  {../../phase2-experiment/results/analysis_outputs/matched_operating_points/}
  {../../phase2-experiment/results/analysis_outputs/support_evidence_chain/}
  {../../phase2-experiment/results/analysis_outputs/cross_family_replication/}
}
\input{macros}

\title{When Top-$K$ Misses the Decision: Tool-Call Drift in Multi-Teacher On-Policy Distillation}

\author{
Jiabin Shen$^{1}$, Guang Chen$^{2}$, and Chengjun Mao$^{3}$\\
$^{1}$\texttt{tjusjb@gmail.com}\\
$^{2}$\texttt{cg234573@antgroup.com}\\
$^{3}$\texttt{chengjun.mcj@antgroup.com}
}

\date{}
\hypersetup{
  pdftitle={When Top-K Misses the Decision: Tool-Call Drift in Multi-Teacher On-Policy Distillation},
  pdfauthor={Jiabin Shen, Guang Chen, Chengjun Mao}
}

\begin{document}
\maketitle

\input{sections/01_abstract}
\input{sections/02_introduction}
\input{sections/03_setup}
\input{sections/04_diagnosis}
\input{sections/05_method}
\input{sections/06_experiments}
\FloatBarrier
\input{sections/07_related_limitations}
\input{sections/08_conclusion}

\clearpage
\bibliographystyle{plainnat}
\bibliography{references}

\clearpage
\appendix
\input{sections/09_appendix}

\end{document}

%% file: macros.tex
\newcommand{\method}{Soft Clamp}
\newcommand{\supportunion}{support union}
\newcommand{\Supportunion}{Support union}

\newcommand{\tooltok}{\texttt{<tool\_call>}}
\newcommand{\llamaentry}{\texttt{\char123\char34}}
\newcommand{\llamaprefix}{\texttt{\char123\char34 name}}

%% file: sections/01_abstract.tex
\begin{abstract}
Teacher top-$K$ logits can preserve nearly all probability mass while omitting
the correction needed for a student's behavioral decision. We study this
failure in routed, two-teacher tool-use distillation. In Qwen3.5-9B, the tool
teacher reinforces the entry coordinate, while response-teacher top-32
retains 99.99\% mass yet contains that coordinate on only 0.4\% of 500
response prompts. A frozen full-vocabulary audit shows that appending its
exact teacher logit nearly recovers full-vocabulary descent on that coordinate.
Forced-entry replay and matched training connect the omission to behavior:
first-position restoration relocates calls to later tokens, whereas
all-position restoration lowers full-generation over-calling from
14.2$\pm$2.1\% to 3.7$\pm$0.5\% across three seeds, while lowering
required-call recall by 12.4 points. Teacher-probability-matched and
output-logit-gradient-rerouted non-tool controls produce smaller shifts and do
not reproduce exact restoration. These findings motivate a second question:
what can different intervention layers change, and at what cost? We compare
student-aware support construction, loss shaping, and validation-tuned
decoding bias. The comparison separates call-threshold movement from modest
gains in discrimination, and relates error costs and system access to
required-call retention and multi-turn success. Llama-3.1-8B reproduces the
directional support asymmetry under native JSON. The results identify
decision-critical support omission as a causal contributor in the primary
Qwen setting and show why preserving teacher mass alone cannot certify
behavioral fidelity.
Code and aggregate artifacts are available in our
\href{https://github.com/shen-jiabin/decision-support-opd}{public repository}.
\end{abstract}

%% file: sections/02_introduction.tex
\section{Introduction}

On-policy distillation (OPD) queries a teacher on student-generated
trajectories and provides dense token-level supervision
\citep{agarwal2024gkd}. Returning full-vocabulary logits at every rollout
position increases teacher output, network transfer, and student-side
processing; high-dimensional logits are also a communication bottleneck in
distributed distillation
\citep{anshumann2025sparse,zhang2025communication}. Our teacher API
therefore returns the teacher's top-$K$ logits. Probability mass makes this
compression appear safe: when top-$K$ retains nearly all teacher mass, the
omitted tail appears unimportant.

Teacher-only top-$K$ preserves tokens likely under the teacher, whereas
distillation error is teacher--student relative. A token can have negligible
teacher probability and still require a strong negative correction when the
student overestimates it. If that token selects a generation mode, such as
\tooltok{} versus a direct response, omitting its correction can affect the
entire continuation. We call this failure \emph{decision-critical support
omission}: compressed teacher supervision preserves probability mass while
removing a coordinate needed to correct the student's behavioral decision.

Multi-teacher tool use makes the asymmetry visible. A tool-call teacher
specializes in structured function calls, a response teacher in direct
answers, and each example is routed to the corresponding teacher
\citep{ma2026mopd}. Under teacher-only top-$K$, the tool route can expose and
reinforce the entry coordinate while the response route omits it. The shared
student then receives directionally incomplete supervision even when routing
is correct. In our primary Qwen3.5-9B setting, vanilla generalized knowledge
distillation (GKD) reaches 91.5\% required-call recall but also 14.2\%
over-calling, compared descriptively with 75.5\% and 4.9\% for a separately
trained mixed-SFT reference. The call-heavy operating point persists in
multi-turn interaction as more calls and tool-call loops.

We organize the study around two linked questions. \emph{First, does omitted
support contribute causally to this behavioral drift? Second, what can
interventions at different layers change, and at what cost?} The first
question requires tracing a coordinate from the routed objective to complete
generations. The second requires distinguishing correction of training
support, movement of the call threshold, and preservation of useful tool
behavior.

For the diagnosis, we first audit full-vocabulary teacher and student logits
on both Qwen routes. On 500 response prompts, teacher top-32 retains 99.99\%
probability mass but contains \tooltok{} on only 0.4\% of prompts. On 500 tool
prompts, the tool teacher ranks it first every time and supplies positive
entry descent in all 1,500 prompt--student pairs. The tool-side direction is
preserved even though its gradient magnitude is much smaller than under full
vocabulary. The bilateral audit therefore separates retained mass, coordinate
coverage, and gradient fidelity. Appending the exact response-teacher logit
nearly recovers full-vocabulary descent on the omitted coordinate.

We then connect this local discrepancy to the trajectory and to training.
Paired forced-entry replay shows that choosing the wrong mode expands
teacher--student divergence over the continuation. Matched first-position
restoration sharply lowers first-token entry error, but calls migrate to
later positions and full-generation over-calling changes little. Restoring
\tooltok{} at every supervised response position closes this temporal escape
route: over-calling falls from 14.2$\pm$2.1\% to 3.7$\pm$0.5\% across three
seeds. A teacher-probability-matched non-tool placebo changes over-calling by
only 0.95 points. A synthetic control that reroutes the selected
output-logit-gradient coefficient to a non-tool coordinate also produces a
smaller shift. These controls narrow explanations based on support expansion
and selected logit-gradient magnitude, while leaving output-row and
full-parameter-gradient equivalence unclaimed.

The diagnostic intervention also reveals why finding a cause does not settle
the intervention choice. All-position restoration lowers required-call
recall to 79.1\% and reduces multi-turn dialogue exact success. Restoring the
omitted correction changes a shared model, so its effects extend beyond
should-respond examples. The downstream study therefore compares three
layers with different correction scopes and access requirements.
Teacher--student support union changes the coordinates used for distillation;
Hard Clip, Global Reweight, and \method{} reshape retained training signals;
validation-tuned entry bias changes decoding without retraining. The union is
a familiar student-aware construction \citep{wang2026teachability}, used here
as a support-level representative rather than a new algorithm.

We evaluate these layers along three axes. First-token scores distinguish
changes in decision separation from movement of the operating point. A
validation-to-test comparison and error-cost sweep show when a decoding
adjustment can approach a trained strategy and when their ordering changes.
Multi-turn evaluation places fewer calls and loops beside required-call
coverage and dialogue success. This comparison exposes both behavioral and
system costs: support union uses wider teacher queries and additional
training time, loss shaping reuses the original top-$K$ interface but requires
retraining, and entry bias requires decoder control. The choice depends on
which errors matter and which part of the system can be changed.

We also test the pattern with Llama-3.1-8B-Instruct under native JSON tool
serialization. Its response teacher retains 99.977\% probability mass at
top-32 while omitting the JSON-entry token on all 500 audited response prompts,
even through $K=256$; the tool teacher ranks the same token first on every
audited tool prompt. Student-aware support nearly recovers full-vocabulary
corrective descent, and matched loss- and support-level training produces
distinct restraint--capability shifts. This extends the support and
intervention evidence across model families and tokenizers; the complete
matched restoration-and-control causal chain remains scoped to Qwen.

Our contributions are:
\begin{itemize}[leftmargin=*]
    \item \textbf{A decision-level failure mode in routed OPD.} A bilateral
    audit shows how near-complete teacher mass can coexist with omitted
    response-side correction and attenuated tool-side gradient magnitude.
    \item \textbf{A causal chain from support to behavior.} Frozen-gradient
    audits, forced-entry replay, temporal-scope restoration, and non-tool
    controls connect the omitted coordinate to entry, delayed calls, and
    full-generation behavior in Qwen. Llama independently reproduces the
    directional support asymmetry and matched intervention pattern.
    \item \textbf{A cross-layer intervention study.} We compare support
    construction, loss shaping, and decoding bias through decision separation,
    operating points, interaction outcomes, and system requirements. The
    comparison explains why correcting an identified omission and selecting a
    useful deployment policy require different evidence.
\end{itemize}

Figure~\ref{fig:support-evidence-chain} summarizes the diagnostic chain;
Sections~\ref{sec:intervention-layers}--\ref{sec:experiments} develop and
compare responses to it. Together, the results motivate evaluating compressed
distillation by the behavioral corrections it preserves and the costs of
recovering them.

%% file: sections/03_setup.tex
\section{Problem Setup}
\label{sec:setup}

\subsection{Multi-teacher on-policy distillation}

Let $x$ be a dialogue context and $y=(y_1,\ldots,y_T)$ a student-generated continuation. On-policy distillation evaluates teacher distributions on this student trajectory. A tag $m \in \{\mathrm{tool}, \mathrm{response}\}$ routes each example to the corresponding teacher.
The tag is teacher-routing metadata and is not included in the student's
input; evaluation does not supply a route oracle.

We use GKD-style token losses based on the divergence between teacher and student next-token distributions \citep{agarwal2024gkd}. Let $p_t(\cdot \mid x,y_{<i})$ be the teacher distribution, $p_s(\cdot \mid x,y_{<i})$ the student distribution, and $S_i^{(K)}$ the teacher's top-$K$ token set. Write $\mathcal{R}_{S}[p]$ for the normalized restriction of $p$ to $S$. The teacher API returns $K=32$ tokens, and we compute
\begin{equation}
    d_i^{(K)} = \mathrm{JSD}\!\left(
    \mathcal{R}_{S_i^{(K)}}[p_t] \,\|\,
    \mathcal{R}_{S_i^{(K)}}[p_s]\right).
\end{equation}
Unless stated otherwise, $p_t$ and $p_s$ are temperature-scaled with
$\tau=0.9$ before support restriction and renormalization. Student rollouts
use the same temperature with top-$p=1$ and no top-$k$ sampling cutoff.
Reported retained-mass diagnostics use the raw, pre-temperature teacher
distribution; all JSD and descent quantities use $\tau=0.9$.
This support-conditioned objective has an exact local consequence: for a student logit $z_{i,v}$ whose token $v\notin S_i^{(K)}$, $\partial d_i^{(K)}/\partial z_{i,v}=0$.
The sequence loss averages supervised target positions. With probability $\lambda=0.8$, the continuation comes from the current student; otherwise, training uses the dataset trajectory. Dataset-source trajectories also receive a supervised format anchor:
\begin{equation}
    \mathcal{L} =
    \mathcal{L}_{\mathrm{distill}}
    + \alpha\,\mathbf{1}[z=\mathrm{dataset}]\,\mathcal{L}_{\mathrm{SFT}},
\end{equation}
where $z$ is the trajectory source and $\alpha=0.3$. This anchor preserves the structured-call protocol and remains fixed across training-time strategies.

\subsection{Tool-call versus response supervision}

Both teachers start from the same base model. One is supervised on structured tool-call targets and the other on direct-response targets. Training supervises the final assistant turn and retains previous turns as context.

Our backend consumes the final target as text, so we pre-render structured calls in the model chat template's tool-call form. Appendix~\ref{app:tool-call-rendering} gives an example.

\subsection{Empirical setting}

Our primary setting uses a Qwen3.5-9B student and two specialized 9B teachers
on APIGen-MT-derived data. A conversation-disjoint decision test separates
should-call from should-respond behavior; a local first-token protocol (E1)
probes entry, while BFCL, When2Call, and a fixed-harness BFCL multi-turn
diagnostic test transfer and interaction. Section~\ref{sec:experiments}
defines these protocols, and Appendices A--B give data and metric details.
Throughout, first-token response-entry error remains distinct from
full-generation over-calling. E1 counts a call when the top-1 first token is
the entry token; its AUC uses the student's entry probability as the score,
with should-call examples as the positive class. Complete-generation metrics
instead detect calls anywhere in the output.

For cross-family replication, we train an independent Llama-3.1-8B-Instruct
student and same-family teacher pair. Qwen uses the single special token
\tooltok{} to enter tool mode. Llama emits native JSON beginning with
\llamaprefix{}; the behavior-switch coordinate audited below is its first
token, \llamaentry{}. Both settings use teacher top-32 supervision and the
same routed tool-call/response construction.

\subsection{Decision-critical support and trajectory leverage}

We call an omitted token \emph{decision-critical} when the teacher assigns it
little probability, the student assigns it enough probability to change a
discrete behavior, and the full-vocabulary objective would directly correct
its logit. Probability-mass coverage need not preserve this property: a set
$S_i^{(K)}$ can retain almost all of $p_t$ while assigning exactly zero direct
gradient to an omitted student error.

\emph{Trajectory leverage} describes how strongly such a token constrains the
continuation. Mode-entry tokens such as \tooltok{}, function names, and
structural markers can select an entire tool-call trajectory; ordinary
response tokens often affect a local phrase. In our routed setting, the tool
teacher exposes \tooltok{} and supplies a positive entry update on every
audited tool prompt, whereas the response teacher usually omits the same token
and loses the corresponding negative update.

This yields a testable chain: truncated support removes a boundary correction,
mode entry selects a trajectory, and mismatch distributes divergence over
later positions. Sections~\ref{sec:mode-mismatch-support}--\ref{sec:support-injection-training}
measure both routed directions, their full-vocabulary magnitudes, and the
trajectory link, then test their integrated effect with matched training.
Section~\ref{sec:alternative-explanations} compares controls for support size,
teacher-probability scale, and selected output-logit-gradient magnitude.

%% file: sections/04_diagnosis.tex
\section{Diagnosing Decision-Support Omission}
\label{sec:diagnosis}

The diagnosis tests three links: whether truncation removes a corrective
coordinate, whether the associated mode choice changes the continuation, and
whether restoring that coordinate changes learned behavior. Frozen audits,
paired forced-entry replay, and matched training address these links in turn.
We then compare alternative explanations before stating the scope of the
causal conclusion.

\subsection{Routed \texorpdfstring{top-$K$}{top-K} support is directionally asymmetric}
\label{sec:mode-mismatch-support}

We recompute the first supervised position with local full-vocabulary teacher
and matched-student logits on both Qwen routes. Each route uses 500 fixed
validation prompts, and student-dependent quantities pair them with vanilla
student seeds 42, 44, and 60, yielding 1,500 teacher--student pairs per route.

On response prompts, the teacher assigns \tooltok{} mean probability only
$6.14\times10^{-8}$. Its top-32 retains 0.999900 raw teacher mass yet contains
the coordinate on only 0.4\% of prompts; even top-256 covers only 52.2\%.
On the complementary tool prompts, the tool teacher ranks \tooltok{} first
every time, includes it in top-32, and produces positive entry descent in all
1,500 matched pairs. Table~\ref{tab:support-sensitivity} reports the bilateral
audit. Entry-logit descent is $-\partial d/\partial z_{\mathrm{tool}}$, so
negative values lower the tool-entry logit and positive values raise it.

\begin{table}[!htbp]
\centering
\small
\setlength{\tabcolsep}{4.5pt}
\begin{tabular}{lrrr}
\toprule
Support & Entry coverage & Raw teacher mass & Entry-logit descent \\
\midrule
Response $K=32$ & 0.4\% & 0.999900 & $-0.000002$ \\
Response $K=64$ & 4.6\% & 0.999959 & $-0.001726$ \\
Response $K=128$ & 22.6\% & 0.999978 & $-0.009159$ \\
Response $K=256$ & 52.2\% & 0.999987 & $-0.021794$ \\
Response T$\cup$S top-32 & 100.0\% & 0.999911 & $-0.040748$ \\
Response $K=32$ + \tooltok{} & 100.0\% & -- & $-0.040756$ \\
Response full vocabulary & 100.0\% & 1.000000 & $-0.040708$ \\
\midrule
Tool $K=32$ & 100.0\% & 1.000000 & $+0.001167$ \\
Tool $K=64$ & 100.0\% & 1.000000 & $+0.003882$ \\
Tool $K=128$ & 100.0\% & 1.000000 & $+0.006175$ \\
Tool $K=256$ & 100.0\% & 1.000000 & $+0.008982$ \\
Tool T$\cup$S top-32 & 100.0\% & 1.000000 & $+0.020065$ \\
Tool full vocabulary & 100.0\% & 1.000000 & $+0.020174$ \\
\bottomrule
\end{tabular}
\caption{Frozen routed Qwen support audit. Coverage and raw pre-temperature
teacher mass use 500 unique prompts per route; descent and union quantities
pool 1,500 matched teacher--student pairs per route and use the training
temperature $\tau=0.9$. Teacher mass neither certifies response-side coordinate
coverage nor tool-side gradient magnitude.}
\label{tab:support-sensitivity}
\end{table}

The response teacher's strict competition rank (one plus the number of
strictly larger logits) places \tooltok{} at median 236.5, with 90th and 99th
percentiles of 614.1 and 1086.0. Increasing teacher-only $K$ therefore improves
coverage but preserves the same teacher-centric ordering: $K=128$ recovers
only 22.6\% of the full-vocabulary mean descent, and $K=256$ still omits
\tooltok{} on 47.8\% of response prompts. Teacher--student top-32 union instead
contains the coordinate in every response-side pair, averages 38.46 support
tokens, and nearly matches the full-vocabulary descent without hard-coding the
coordinate.

The tool route separates direction from magnitude. Top-32 preserves the
reinforcing direction but substantially attenuates its full-vocabulary
magnitude ($+0.001167$ versus $+0.020174$), despite retained teacher mass
displayed as 1.000000. Teacher--student union averages 58.16 coordinates on
this route and reaches $+0.020065$. Thus mass can hide a missing coordinate on
one route and a weak student-relative update on the other. The routed
directional-asymmetry claim depends on the sign; the magnitude gap further
shows why mass is not a gradient-fidelity diagnostic.
Appendix~\ref{app:mode-mismatch-support} gives response-side per-seed descents,
ranks, and replay details.

\FloatBarrier
\subsection{Wrong mode entry amplifies along the trajectory}

\paragraph{Sequence aggregates hide the boundary mechanism.}
Aggregate statistics yield no stable direction of teacher dominance. Across
31 paired diagnostic steps per method, tool/response point estimates are
0.81--0.93 for token exposure, 0.95--1.03 for raw per-token JSD, and
1.19--1.34 for the squared-divergence proxy; every paired 95\% bootstrap
interval includes one. Yet the largest 1\% of supervised tokens carry
41.2$\pm$0.4\% of diagnostic-sample JSD in vanilla GKD. These sequence
averages locate concentration but neither expose the routed sign asymmetry nor
identify its behavioral leverage. Appendices~\ref{app:aggregate-sanity}
and~\ref{app:token-concentration} give the full audits.

We replay vanilla GKD checkpoint-319 on 500 tool-tagged and 500
response-tagged APIGen-MT validation prompts for each of three student seeds.
Only 5.3\% of tool-tagged rollouts answer directly, but their per-token JSD is
55.45 times that of aligned tool rollouts (95\% CI [41.74, 76.44]). These 80
examples contribute 55.5\% of the squared-divergence proxy over all 3,000
replays. The reverse response$\rightarrow$tool mismatch is more common
(11.6\%) but only 2.03 times its aligned baseline [1.71, 2.39].

Natural mismatch may select harder prompts, so we also force either a tool or
response entry on the same 200 prompts per teacher tag, then let the same
seed-42 student continue. The mismatch/alignment JSD ratio remains 26.73 under
the tool teacher [14.87, 67.91] and 2.67 under the response teacher [2.29,
3.14]. The branches share the pre-entry distribution; divergence opens after
the mode choice and persists over later tokens. Entry is the trigger, not the
sole high-loss position.

\FloatBarrier
\subsection{Support scope controls entry, delayed calls, and final behavior}
\label{sec:support-injection-training}

We turn the frozen audit into two matched training interventions. Both append
\tooltok{} with its exact response-teacher logit whenever top-32 omits it, and
leave tool-tagged examples unchanged. \emph{First-position support} acts only
at the first effective supervised response position. \emph{All-position
support} acts at every supervised response position. Each intervention uses
seeds 42, 44, and 60 with the same initialization, data order, teachers,
319-step budget, and anchored GKD recipe as vanilla.

First-position support verifies the local gradient but leaves an escape route.
Response entry error falls from 14.47$\pm$1.53\% to 3.40$\pm$0.26\%, while
tool-entry call recall also falls from 91.72$\pm$1.22\% to
80.57$\pm$0.68\%. Boundary AUC changes only from 0.9692 to 0.9712. In complete generations,
first-position calls on should-respond examples fall from 13.42\% to 3.07\%,
but delayed calls rise from 0.78\% to 10.42\%; over-calling therefore changes
only from 14.20$\pm$2.08\% to 13.48$\pm$1.27\%.

\begin{figure}[!htbp]
\centering
\includegraphics[width=\linewidth]{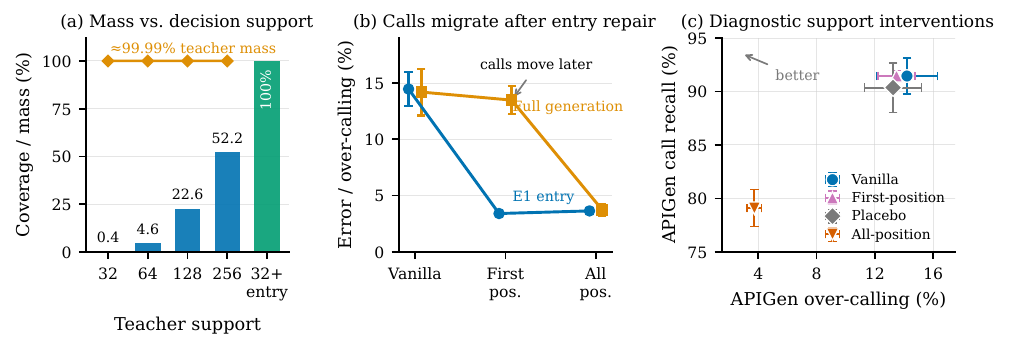}
\caption{Evidence chain under teacher-top-32. (a) The response teacher retains
approximately 99.99\% probability mass at $K=32$ and more at larger $K$, while
\tooltok{} coverage rises from 0.4\% at $K=32$ to only 52.2\% at $K=256$;
appending the exact entry coordinate yields 100\% coverage. (b) First-position
restoration lowers E1 entry error but leaves full-generation over-calling near
vanilla because calls migrate later; all-position restoration closes this
temporal escape route. (c) On APIGen, first-position restoration and the
matched placebo remain near vanilla in full generation, whereas all-position
restoration moves both over-calling and required-call recall. Panel (a)'s teacher-only mass and coverage
curves use 500 unique response prompts; gradient quantities pair them with
three matched students. Error bars in (b)--(c) are sample standard deviations over matched seeds 42,
44, and 60.}
\label{fig:support-evidence-chain}
\end{figure}

All-position support leaves this first-token operating point nearly unchanged:
response entry error is 3.63$\pm$0.24\%, call entry is
81.30$\pm$0.51\%, and AUC is 0.9725$\pm$0.0010. It instead closes the temporal
escape route. Delayed calls on should-respond examples fall to 0.65\%, and
no-call outputs rise to 96.27\%. Full-generation over-calling reaches
3.73$\pm$0.51\%, with reductions of 10.10, 8.35, and 10.80 points relative to
first-position support in the three matched seeds.
Figure~\ref{fig:support-evidence-chain} links the frozen audit to the temporal
scope and final operating points.

The stronger intervention is deliberately diagnostic, not a recommended
training recipe. Relative to first-position support, all-position support
lowers call recall by 12.40 points and BFCL tool-call quality by 2.31 points.
In the multi-turn harness, empty-ground-truth no-call accuracy rises from
47.41\% to 61.49\%, while observed required-turn call coverage falls from
83.15\% to 73.24\%. On the three categories fully implemented by the harness,
turn-level protocol success falls from 32.49\% to 28.60\%. This bidirectional
effect matters: restoring the omitted coordinate
controls final over-calling, but shared parameters do not confine the update to
response examples.

Appendix~\ref{app:support-injection-training} reports call positions, per-seed
deltas, and paired multi-turn intervals.

\FloatBarrier
\subsection{Alternative explanations and identification scope}
\label{sec:alternative-explanations}

Exact restoration changes both the support and the update assigned to the
entry coordinate. Table~\ref{tab:diagnostic-controls-main} places its
behavioral effects beside two non-tool controls. The first matches support
size and teacher-probability scale; the second additionally matches the
selected output-logit-gradient magnitude in a frozen comparison.

\begin{table}[!htbp]
\centering
\small
\setlength{\tabcolsep}{4pt}
\begin{tabular}{@{}lrrrrr@{}}
\toprule
& \multicolumn{2}{c}{E1: first token}
& \multicolumn{2}{c}{APIGen: full generation} & BFCL multi-turn \\
\cmidrule(lr){2-3}\cmidrule(lr){4-5}\cmidrule(l){6-6}
Treatment & Resp. error & Call recall & Over-call & Call recall & Dialogue exact \\
\midrule
Vanilla GKD & 14.47$\pm$1.53 & 91.72$\pm$1.22 & 14.20$\pm$2.08 & 91.45$\pm$1.68 & 7.04$\pm$0.31 \\
First-position support & 3.40$\pm$0.26 & 80.57$\pm$0.68 & 13.48$\pm$1.27 & 91.47$\pm$0.50 & 5.83$\pm$1.01 \\
All-position support & 3.63$\pm$0.24 & 81.30$\pm$0.51 & 3.73$\pm$0.51 & 79.07$\pm$1.73 & 3.38$\pm$0.50 \\
\midrule
Probability placebo & 13.37$\pm$1.10 & 91.30$\pm$1.00 & 13.25$\pm$1.96 & 90.37$\pm$2.31 & 4.92$\pm$0.40 \\
Gradient reroute & 9.25$\pm$3.43 & 85.83$\pm$2.92 & 11.65$\pm$2.88 & 86.93$\pm$3.31 & 5.21$\pm$0.44 \\
\bottomrule
\end{tabular}
\caption{Matched diagnostic interventions (percent; mean$\pm$sample std over
seeds 42, 44, and 60). All share the 319-step training budget and anchored
recipe. E1 entry and full-generation decisions are distinct measurements.
Dialogue exact is the local multi-turn execution/state endpoint defined in
Section~\ref{sec:experiments}. Both non-tool controls use all-position response
scope; their matching conditions and limits are described below.}
\label{tab:diagnostic-controls-main}
\end{table}

\paragraph{Matched support expansion does not explain the effect.}
All-position restoration changes both support cardinality and token identity.
We therefore train a probability-matched placebo over the same three seeds.
At each supervised response position, it appends the non-tool tail token whose
teacher probability is closest to that of \tooltok{}, matching scope and
probability scale without restoring the decision coordinate. Placebo
over-calling is 13.25$\pm$1.96\%, only 0.95 points below vanilla, and its E1
response-entry error is 13.37$\pm$1.10\%, only 1.10 points lower. Exact
all-position restoration reduces the two metrics by 10.47 and 10.83 points.
The placebo is not behaviorally inert---overall dialogue exact success falls
to 4.92$\pm$0.40\%---but it does not reproduce the target-token effect. It
therefore rules out a generic support-cardinality explanation at matched
teacher-probability scale, without matching student probability or effective
gradient strength. Appendix~\ref{app:support-construction-controls} reports the
full comparison.

\paragraph{Matched output-logit magnitude alone does not explain the effect.}
The synthetic output-logit-gradient-reroute control uses the placebo's legal
non-tool candidate but preserves exact restoration's frozen forward loss and
selected output-logit gradient coefficient and norm. It then routes that
coefficient to the candidate output row. Despite this match, E1 response error
and APIGen over-calling remain 9.25$\pm$3.43\% and 11.65$\pm$2.88\%, versus
3.63$\pm$0.24\% and 3.73$\pm$0.51\% under exact restoration. The per-seed
over-calling gaps are $+8.25$, $+4.90$, and $+10.60$ points. Thus matching the
selected output-logit magnitude on another coordinate does not reproduce
restoration of the identified coordinate. This control does not match the
output row or full parameter gradients: a coefficient $g$ contributes
$gW_{\mathrm{tool}}$ to the hidden-state gradient under exact restoration,
but $gW_{\mathrm{candidate}}$ under rerouting. Its When2Call score also falls
to 26.70$\pm$22.32\%, from vanilla's 65.57$\pm$1.07\%.
Appendix~\ref{app:gradient-reroute} gives the straight-through formula, frozen
audit, and full benchmark results.

\paragraph{Direct suppression is a separate, limited control.}
An additional three-seed experiment adds a direct penalty on student
tool-entry probability at every supervised response position, using the
original top-32 distillation support. At the fixed weight 0.15, it reduces
full-generation over-calling from 14.20$\pm$2.08\% to
12.93$\pm$1.14\%, while call recall changes from 91.45$\pm$1.68\% to
90.40$\pm$2.05\% and dialogue exact success falls from
7.04$\pm$0.31\% to 4.96$\pm$0.07\%. This setting shows the effect of another
way to suppress the same token; it does not establish the performance of a
tuned penalty or match exact restoration's gradient.
Appendix~\ref{app:support-construction-controls} specifies the penalty.

\paragraph{What the evidence establishes.}
Frozen coordinate recovery, forced-entry replay, temporal-scope restoration,
and the non-tool controls jointly implicate decision-critical support
omission as one causal contributor in the primary Qwen setting. The training
effect depends on both the corrected coordinate and where it is supervised;
the first-token boundary alone does not predict complete-generation behavior.
These experiments do not establish equivalence to full-vocabulary OPD
training, or that the coordinate fully mediates the drift. The accompanying
loss of required calls makes the distinction between identifying a cause and
choosing an intervention consequential.

\FloatBarrier
\subsection{Cross-family support replication}
\label{sec:cross-family-support}

We repeat the frozen first-position audit with Llama-3.1-8B's response teacher
and native JSON-entry token \llamaentry{}. Over 500 unique response prompts,
top-32 retains 0.999770 teacher mass but never contains the entry token; its
coverage remains zero through $K=256$, despite 0.999968 retained mass. The
median entry rank is 105,952.5. Pairing the same prompts with three vanilla
students yields 1,500 teacher--student pairs: the teacher/student top-32 union
contains the entry token for 97.93\%, averages 40.49 coordinates, and produces
mean entry descent $-0.04923$, nearly identical to the full-vocabulary
$-0.04918$. A complementary audit over 500 unique tool prompts finds the
opposite side of the asymmetry: the tool teacher ranks the JSON-entry token
first on every prompt, assigns it mean probability 0.999999, and includes it
in top-32 with mean entry descent $+0.00517$ across the 1,500 matched
teacher--student pairs; the full-vocabulary descent is $+0.00903$. Thus
teacher-top-32 explicitly reinforces entry on the tool route but supplies no
direct response-route correction. Qwen provides the matched restoration
experiments; this frozen Llama audit, together with the matched support-union
intervention in Section~\ref{sec:robustness}, shows that the support blind spot and
its student-aware correction survive a different tokenizer and tool
serialization.
Appendix~\ref{app:llama-replication} gives the full sweep and per-seed scope.

\paragraph{From diagnosis to intervention choice.}
Restoration identifies a missing correction, but its strongest behavioral
effect also suppresses required calls. The next question is therefore what
different intervention layers can change, and which costs accompany those
changes. Section~\ref{sec:intervention-layers} defines support construction,
loss shaping, and decoding bias under common comparison axes.
Section~\ref{sec:experiments} distinguishes their effects on decision
separation, operating points, and interaction success, then evaluates how
error costs and available system access change the deployment preference.

%% file: sections/05_method.tex
\section{Cross-Layer Optimization Strategies}
\label{sec:intervention-layers}

The intervention study asks which parts of the diagnosed drift can be changed
at different layers, and what resources and capabilities each change costs.
Support construction changes the coordinates being distilled. Loss shaping
changes the strength of retained training signals. Decoding bias changes the
deployed entry threshold. These mechanisms motivate separate measurements of
support recovery, decision separation, operating points, and task success.
Exact restoration and its matched controls retain their diagnostic role.

\subsection{Roles and comparison axes}

All training-time interventions use the same teacher pair, data, OPD settings,
and format anchor. Diagnostic restoration, its matched controls, and
the student-aware support strategy alter support; the generic loss shapers
alter token-level divergence; entry bias changes the first tool-entry logit at
inference. The dataset-source format anchor from Section~\ref{sec:setup}
remains fixed. Table~\ref{tab:calibration-taxonomy} makes these roles and
access requirements explicit.

\begin{table}[!ht]
\centering
\small
\setlength{\tabcolsep}{4pt}
\begin{tabularx}{\linewidth}{@{}l l X c X@{}}
\toprule
Strategy & Role / layer & Correction scope & Retrain & Access requirement \\
\midrule
Vanilla GKD & Reference & Teacher top-$K$ objective & Yes & Original top-$K$ interface \\
Exact restoration & Diagnostic & Known response coordinate & Yes & Exact teacher coordinate \\
Probability placebo & Diagnostic & Matched non-tool coordinate & Yes & Matched teacher coordinate \\
Gradient reroute & Diagnostic & Matched non-tool logit gradient & Yes & Matched teacher coordinate \\
Entry penalty & Diagnostic & Direct response-token suppression & Yes & Original top-$K$ interface \\
\Supportunion{} & Support & Distilled token coordinates & Yes & Student-selected teacher probes \\
Hard Clip & Loss & Retained loss above fixed cap & Yes & Original top-$K$ interface \\
Global Reweight & Loss & Retained batch-relative loss & Yes & Original top-$K$ interface \\
\method{} & Loss & Retained extreme gradients & Yes & Original top-$K$ interface \\
Entry bias & Decoding & Inference entry threshold & No & Validation data; decoder control \\
\bottomrule
\end{tabularx}
\caption{Cross-layer strategy taxonomy. Exact restoration and its matched
controls are diagnostic interventions. Among the evaluated optimization layers,
only support-level construction places student-selected omitted coordinates
into the training support; loss shaping acts on retained signals, and entry
bias acts during decoding.}
\label{tab:calibration-taxonomy}
\end{table}
\FloatBarrier

\subsection{Support-level optimization}

The support-level representative uses the teacher/student top-$K$ union as
the actual within-position support:
\begin{equation}
U_i^{(K)}=\operatorname{TopK}(p_t)_i\cup\operatorname{TopK}(p_s)_i.
\end{equation}
We query teacher logits for student-selected tokens missing from teacher
top-$K$ and evaluate the same restricted JSD on $U_i^{(K)}$. This familiar
construction \citep{wang2026teachability} can recover student-relevant
coordinates without hard-coding \tooltok{} or its position. Here it changes
within-position support; Teachability-Aware OPD uses a union to score
sequence-position compatibility.

Table~\ref{tab:support-sensitivity} establishes its frozen coordinate recovery
on both Qwen routes. Its training cost is a separate quantity: over the 319
matched steps, realized support averages 38.88$\pm$0.08 tokens, or 21.5\%
more returned coordinates than teacher top-32. Observed steady-state step
time rises from 106.4 to 123.5 seconds (16.0\%). Thus this comparison holds
the epoch and step budget fixed while allowing additional teacher queries and
computation. Appendix~\ref{app:support-construction-controls} reports overlap,
per-seed results, and paired behavioral deltas.

\subsection{Loss-level optimization}

Loss shaping tests whether changing the concentration of retained training
signals moderates the behavioral shift. Let $d_i$ denote the original
restricted-support divergence and $\operatorname{sg}$ stop gradients. The
three representatives use
\begin{equation}
\begin{aligned}
\text{Hard Clip:}\quad & d'_i=\min(d_i,c),\\
\text{Global Reweight:}\quad & d'_i=d_i\,\operatorname{sg}(w_i),\\
\text{\method{}:}\quad & d'_i=d_i\min\!\left(1,\frac{C}{\operatorname{sg}(d_i)}\right)
    \quad (d_i>0),
\end{aligned}
\end{equation}
with zero divergence left unchanged. Hard Clip uses $c=0.5$ and removes
marginal gradients above the cap. Global Reweight uses normalized, clipped
exponential weights of batch-standardized divergence
\citep{hao2026gndpo}, so it reweights non-extreme tokens as well.
\method{} sets $C=\operatorname{sg}(3\,\operatorname{mean}_i d_i)$ and keeps
a nonzero gradient scaled by $C/d_i$ above the threshold.

All three reuse the original teacher top-$K$ interface and require
retraining. They cannot directly restore an omitted coordinate; behavioral
effects arise through changes to retained signals and shared parameters.
Appendix~\ref{app:objective-details} gives the complete weight definition and
fixed hyperparameters. These settings are held fixed before multi-seed test
evaluation.

\subsection{Decoding-level optimization}

Entry bias moves the deployment operating point without retraining. At the
first assistant token of each generation step, it replaces the tool-entry
logit $\ell_{\text{tool}}$ with $\ell_{\text{tool}}+b$, where $b\le 0$.
We sweep $b\in[-5,0]$ in increments of 0.05 on APIGen-MT validation,
independently for each vanilla seed. Two operating points match either
\method{}'s validation response-entry error or its call recall; ties use the
smallest $|b|$. Each selected bias is frozen for test and strict multi-turn
evaluation. The cost-sensitive analysis separately selects a bias minimizing
validation risk for each specified error-cost ratio
(Section~\ref{sec:cost-main}).

This intervention requires a controllable decoder and validation-time tuning.
It directly changes only the first-token logit, although the resulting entry
choice can alter the continuation. The local decoding protocol therefore
evaluates both entry and interaction effects.

The three layers therefore exchange different resources and capabilities:
support construction requires wider teacher queries, loss shaping retains
the original interface but retrains the model, and entry bias avoids training
but requires decoder control. We compare their effects and costs as responses
to the same diagnosed drift.

%% file: sections/06_experiments.tex
\section{Experiments}
\label{sec:experiments}

We evaluate the intervention layers through three complementary comparisons:
full-generation operating points, first-token separation and error costs, and
multi-turn interaction outcomes. A reduction in over-calling is assessed
alongside required-call retention and task success. This makes the second
research question empirical: the relevant choice depends on which behavior
changes, what is lost, and which system access the strategy requires.

\subsection{Models and training}

The main experiments use Qwen3.5-9B (Qwen/Qwen3.5-9B) \citep{qwen2026qwen35hf}. Tool-call and response teachers start from this checkpoint, train on their respective target types, and remain frozen while GKD queries them on student rollouts. The optimization study includes a support-level strategy (\supportunion{}), three loss-level strategies (Hard Clip, Global Reweight, and \method{}), and a decoding-level strategy (entry bias), with base, Mixed SFT, and vanilla GKD as references. The matched support-scope interventions in Section~\ref{sec:support-injection-training} provide the causal test; the probability-matched placebo and gradient-reroute control narrow alternative explanations. All GKD students share data, teachers, training length, core OPD settings, and the $sft\_alpha=0.3$ format anchor.

Section~\ref{sec:robustness} changes model family, tokenizer, teachers, and
tool serialization with Llama-3.1-8B-Instruct
\citep{meta2024llama31modelcard}. It repeats mixed SFT, vanilla GKD,
\method{}, and the same \supportunion{} intervention as representatives of the
loss and support training layers under $K=32$, $\lambda=0.8$,
$\alpha=0.3$, and three-seed protocol. Its native calls are rendered as
assistant JSON and normalized to the common tool-call representation before
evaluation. Appendix~\ref{app:qwen4b} reports a secondary Qwen3.5-4B
student-scale check of the loss-level strategies; because that experiment does
not include support construction, it is not used as mechanism replication.

Unless marked otherwise, benchmark tables report mean$\pm$std over three
seeds for GKD variants; Base and Mixed SFT are single reference runs. The
multi-turn Hard Clip row and explicitly marked diagnostic ablations use a
representative run.

\subsection{Datasets and metrics}

\paragraph{APIGen-MT-derived decision set.}
APIGen-MT \citep{prabhakar2025apigenmt} supplies our main in-domain decision
set. We construct a conversation-disjoint evaluation from held-out
conversations. After removing thinking blocks, we apply family-specific
parsers and normalize detected calls to one canonical representation: Qwen
uses its XML \tooltok{} block, whereas Llama uses a native JSON object with
\texttt{name} and \texttt{parameters} fields. A generation counts as a
tool-call decision when the corresponding parser detects a call; schema
validity is not required by these APIGen decision metrics, while BFCL evaluates
call quality separately. The balanced test reports decision
accuracy, over-calling, should-call recall, and should-respond recall, making
the call/response trade-off explicit.

\paragraph{BFCL.}
BFCL \citep{patil2025bfcl} evaluates function-calling quality and irrelevance refusal. We use it to test whether tool-call calibration transfers beyond APIGen-style decision points.

\paragraph{When2Call.}
When2Call \citep{ross2025when2call} evaluates whether a model should call a tool, request more information, or refuse. Its multiple-choice formulation provides an out-of-domain decision diagnostic.

\paragraph{BFCL multi-turn loop diagnostic.}
We also run a fixed-harness multi-turn diagnostic on 800 BFCL tasks and 3,136 user turns. It records calls per turn, loop and max-step rates, repeated calls, and non-tool finals to test whether single-turn over-calling persists through interaction.
Loop@3 denotes at least three consecutive tool-call steps within a user turn.
Required-turn call coverage measures whether the model makes a call on
observed turns that require one.
We additionally report overall dialogue exact success: the BFCL execution/state checker
must accept the complete trajectory, and turns with empty ground truth must
emit no executable tool call. This task-level metric is all-or-nothing across
the dialogue. It is a local harness metric, excludes final natural-language
answer quality, and is not an official leaderboard score.
To expose compounding behind its low absolute scale, we also
replay saved calls through the same checker at every observed turn. Our local
harness does not issue BFCL's empty-message function-supplementation step, so
turn-level endpoint summaries mark the missing-function category unsupported
rather than treating its structural zero as model evidence.

\subsection{Training-time APIGen-MT operating points}

The main comparison asks whether an intervention can retain much of vanilla
GKD's call-recall gain while reducing its over-calling side effect.

\begin{table}[!htbp]
\centering
\small
\begin{tabular}{lrrrr}
\toprule
Method & Decision Acc & Over-calling & Call Recall & Respond Recall \\
\midrule
Base & 80.7 & 7.2 & 68.5 & 92.8 \\
Mixed SFT & 85.3 & 4.9 & 75.5 & 95.1 \\
Vanilla GKD & 88.6$\pm$0.3 & 14.2$\pm$2.1 & 91.5$\pm$1.7 & 85.8$\pm$2.1 \\
Hard Clip & 88.7$\pm$1.0 & 11.9$\pm$0.9 & 89.2$\pm$3.0 & 88.2$\pm$0.9 \\
Global Reweight & 88.6$\pm$1.3 & 9.7$\pm$0.8 & 86.9$\pm$3.3 & 90.3$\pm$0.8 \\
\method{} & 88.7$\pm$0.7 & 9.0$\pm$0.2 & 86.5$\pm$1.4 & 91.0$\pm$0.2 \\
\Supportunion{} & 89.8$\pm$0.8 & 7.4$\pm$0.6 & 87.0$\pm$2.0 & 92.6$\pm$0.6 \\
\bottomrule
\end{tabular}
\caption{APIGen-MT-derived decision results. All values are percentages. GKD rows report mean$\pm$std over three seeds; Base and Mixed SFT are single reference runs. Loss shaping and support construction move different call/response operating points.}
\label{tab:apigen}
\end{table}

Table~\ref{tab:apigen} compares the support- and loss-level training
strategies; the decoding-level comparison follows in
Section~\ref{sec:e1-main}. Mixed SFT favors responses, while vanilla GKD
raises call recall to 91.5\% but over-calls on 14.2\% of should-respond
examples. Relative to vanilla's 14.2\% over-calling and 91.5\% call recall,
\method{} moves these metrics to 9.0\% and 86.5\%, while support union moves
them to 7.4\% and 87.0\%. The diagnostic all-position restoration reaches
3.7\% and 79.1\%, illustrating a more aggressive restraint--capability shift
rather than an optimization target. Hard Clip and Global Reweight fill out
the loss-shaping range at 9.7--11.9\% over-calling and 86.9--89.2\% call recall. These
operating points exchange correction scope, teacher access, retraining, and
required-call retention.
Appendix~\ref{app:apigen-extra} gives the trade-off plot and qualitative
examples.

\subsection{Threshold movement and decision separation}
\label{sec:e1-main}

The APIGen results compare fixed model outputs. We next place support-, loss-,
and decoding-level strategies under a common local first-token protocol to
identify whether each layer changes threshold-free separation, the operating
point, or both under its deployment requirements.
With thinking disabled, E1 uses a local Transformers scorer on 4,000 APIGen-MT
test examples. Entry probability is the AUC score and should-call is the
positive class; entry decisions use the top-1 first token. Full-generation
APIGen instead uses vLLM, so absolute rates are compared within each protocol.

At the loss layer, \method{} changes AUC only from
0.9692$\pm$0.0023 to 0.9710$\pm$0.0011, indicating primarily an
operating-point shift. At the decoding layer, a validation-selected
non-positive bias nearly matches its response-entry error without retraining
(11.2$\pm$0.6\% versus 11.1$\pm$0.3\%), but has lower call-entry recall
(89.2$\pm$1.2\% versus 90.4$\pm$0.2\%). These test values use biases selected
to match validation response-entry error, not biases tuned on the test set.
The support-level operating point has
0.9760$\pm$0.0006 AUC, 8.72$\pm$0.74\% response-entry error, and
90.55$\pm$1.00\% call-entry recall, consistent with modest separation change
as well as threshold movement.

\begin{figure}[!htbp]
\centering
\includegraphics[width=0.68\linewidth]{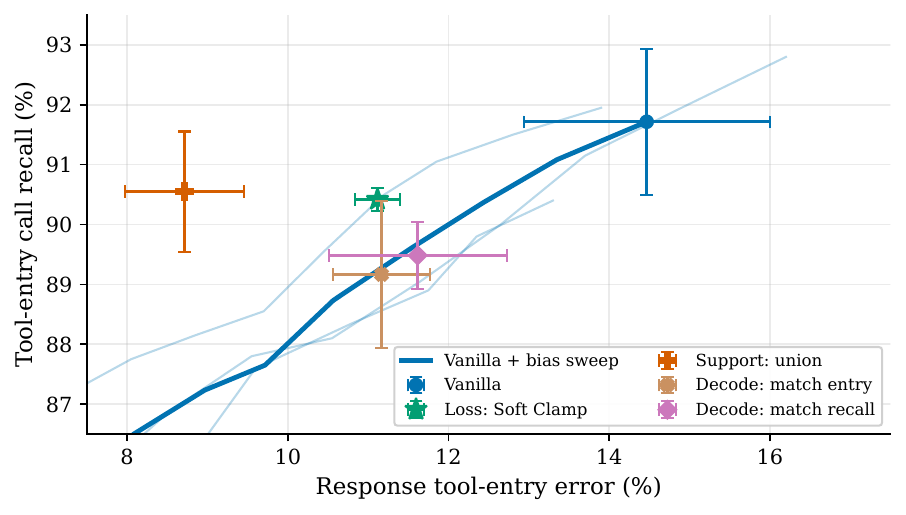}
\caption{E1 first-token operating points on APIGen-MT test. Left and up are
better. Blue lines sweep a \tooltok{} bias for vanilla GKD; the green star is
loss-level \method{}, and the orange plus is the \supportunion{} intervention.
Diamonds are validation-selected decoding biases. Pale lines show individual
vanilla seeds; error bars are sample standard deviations over three seeds.}
\label{fig:e1-boundary-pareto}
\end{figure}

Figure~\ref{fig:e1-boundary-pareto} places the three layers under this common
protocol. The support-union AUC increase over vanilla is 0.68 percentage
points, with a descriptive three-seed paired 95\% $t$ interval of [0.20, 1.15].
This is evidence of a modest change in ranking, alongside the larger shifts in
call rates. Appendix~\ref{app:e1-boundary-bias} reports both validation-matched
bias choices and their per-seed values.

\FloatBarrier

\subsection{Error costs and deployment preferences}
\label{sec:cost-main}

To choose among operating points, we must specify the relative cost of the
two errors. Let $\rho=C_{\mathrm{over}}/C_{\mathrm{miss}}$ be the cost of an
unnecessary call relative to a missed required call. For response-side
false-positive rate $e_{\mathrm{over}}$ and call-side false-negative rate
$e_{\mathrm{miss}}$, we report the class-balanced normalized risk
\begin{equation}
R_\rho=\frac{\rho e_{\mathrm{over}}+e_{\mathrm{miss}}}{\rho+1}.
\label{eq:cost-risk}
\end{equation}
The class-conditional errors are weighted equally before applying the cost
ratio; deployments with other class prevalences require their own weighting.
For each seed and each $\rho$, the E1 cost-aware bias minimizes validation
risk and is then frozen on test. This selection is separate from matching
\method{}'s validation entry error or call recall.

\begin{figure}[!htbp]
\centering
\includegraphics[width=\linewidth]{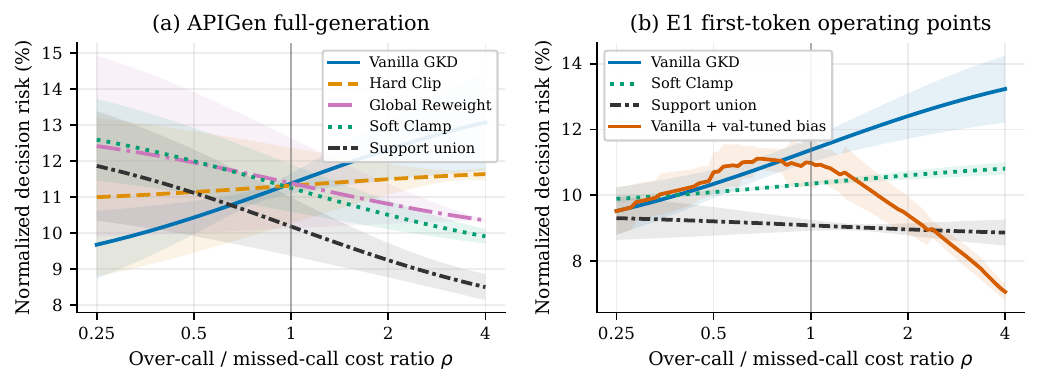}
\caption{Cost-sensitive operating points. Lower is better. Panel (a) compares
fixed trained GKD variants on APIGen full-generation decisions. Panel (b)
compares fixed vanilla GKD, \method{}, and \supportunion{} with a per-seed
inference bias selected on E1 validation for each cost ratio and frozen on
test. Lines and bands are three-seed means and sample standard deviations.
The panels use distinct protocols and are not pooled.}
\label{fig:cost-sensitive-operating-points}
\end{figure}

Figure~\ref{fig:cost-sensitive-operating-points} shows how error costs change
the relative preference. On APIGen full generations, vanilla GKD has the
lowest mean risk at $\rho\in\{0.25,0.5\}$, where missed required calls cost
more, while \supportunion{} is lowest at $\rho\in\{1,2,4\}$. On E1,
\supportunion{} has the lowest mean risk through $\rho=2$ among the evaluated
points; cost-aware bias is lowest at $\rho=4$, where its mean risk is
7.06\% versus 8.86\% for support union. These comparisons are descriptive
over three seeds.

The ordering is conditional on both error costs and access. The support
strategy incurs the additional queries and training time reported in
Section~\ref{sec:intervention-layers}; loss shaping uses the original top-$K$
interface; entry bias requires validation data and decoder control. The risk
sweep evaluates decision errors, while the next comparisons assess transfer
and interaction outcomes that this two-error summary does not capture.
\FloatBarrier

\subsection{BFCL and When2Call}

Single-turn transfer varies modestly across the trained strategies. The three
loss-shaping methods span 79.8--80.8\% BFCL single-turn overall and 64.3--67.1\%
When2Call MCQ, while support union reaches 81.53$\pm$0.93\% and
68.27$\pm$1.11\%. Rankings change across columns
(Appendix~\ref{app:bfcl-when2call}). Mixed SFT remains stronger on BFCL, and
all GKD variants remain below Base/Mixed SFT on When2Call overall, exposing
residual capability limits alongside the operating-point changes.

\subsection{Multi-turn tool-call loops}

The multi-turn diagnostic connects boundary drift to interaction behavior.
Table~\ref{tab:multiturn} places calls and loops beside the required-call and
dialogue endpoint metrics for all evaluated loss shapers and the support
strategy. \method{} and support union reduce calls and loops to different
degrees, with corresponding reductions in required-turn coverage and dialogue
exact success. The loop metric alone would therefore give an incomplete
account of either strategy's utility.

\begin{table}[!htbp]
\centering
\small
\setlength{\tabcolsep}{6pt}
\begin{tabular}{lrrrr}
\toprule
Method & Calls/turn & Loop@3 & Required call & Dialogue exact \\
\midrule
Mixed SFT & 0.974 & 5.1 & 79.77 & 5.12 \\
Vanilla GKD & 1.515$\pm$0.122 & 15.1$\pm$3.0 & 92.90$\pm$0.33 & 7.04$\pm$0.31 \\
Hard Clip & 1.348 & 11.5 & 88.88 & 6.00 \\
Global Reweight & 1.297$\pm$0.088 & 10.8$\pm$1.4 & 87.78$\pm$4.62 & 5.33$\pm$0.92 \\
\method{} & 1.289$\pm$0.023 & 10.3$\pm$0.4 & 88.50$\pm$2.01 & 6.17$\pm$0.38 \\
\Supportunion{} & 1.118$\pm$0.057 & 8.0$\pm$0.9 & 81.22$\pm$2.38 & 4.12$\pm$0.33 \\
\bottomrule
\end{tabular}
\caption{BFCL multi-turn interaction and capability trade-offs.
Values except calls/turn are percentages. Required call denotes call coverage
on observed required turns; dialogue exact checks execution/state and no-call
conditions across the complete task. Multi-seed rows report mean$\pm$sample
std; Mixed SFT and Hard Clip are single runs. Additional loop, termination,
and turn-level correctness metrics are in Appendix~\ref{app:bfcl-multiturn}.}
\label{tab:multiturn}
\end{table}

Under a separate strict local-decoder protocol, validation-tuned entry bias
moves calls per turn from 1.734 to 1.492 and Loop@3 from 20.5\% to 15.7\%
without retraining. These values compare biased and unbiased vanilla within
that decoder; they are not pooled with the vLLM-based results in
Table~\ref{tab:multiturn} (Appendix~\ref{app:e1-boundary-bias}).
Thus the interaction study complements the first-token and cost analyses:
reducing call propensity can reduce loops, while task completion requires
preserving the calls that the interaction needs.
\FloatBarrier

\subsection{Cross-family replication}
\label{sec:robustness}

Llama-3.1-8B tests whether a call-heavy vanilla operating point,
response-side support omission, and distinct matched loss- and support-level
shifts appear under a different model family and native JSON protocol.

\begin{figure}[H]
\centering
\includegraphics[width=\linewidth]{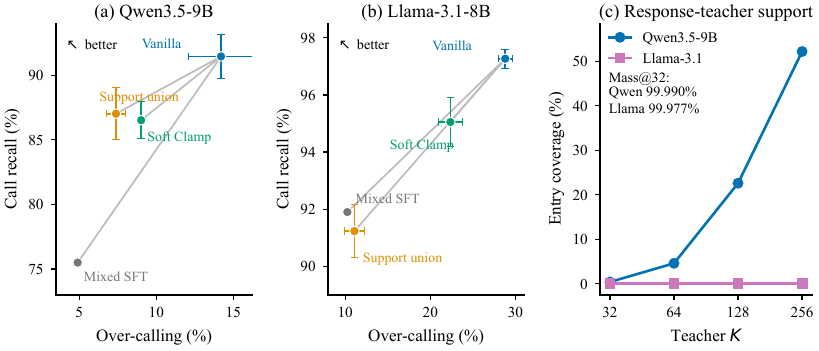}
\caption{Cross-family replication. Panels (a)--(b) show APIGen operating
points; upper-left is better for the displayed call/response trade-off. Gray
segments connect the separate mixed-SFT reference, vanilla GKD, and matched
training interventions; they do not denote a training trajectory. Error bars
are sample standard deviations over three seeds; mixed SFT is a single
reference. Panel (c) audits response-teacher
entry coverage as $K$ grows: Qwen uses \tooltok{} and Llama uses the native
JSON-entry token \llamaentry{}. Their top-32 teacher masses are 99.990\% and
99.977\%, respectively, despite low or zero decision-token coverage.}
\label{fig:cross-family-replication}
\end{figure}

\begin{table}[!htbp]
\centering
\small
\setlength{\tabcolsep}{6pt}
\begin{tabular}{lrrrr}
\toprule
Method & Dec. Acc & Over-call & Call Rec. & Resp. Rec. \\
\midrule
Mixed SFT & 90.83 & 10.25 & 91.90 & 89.75 \\
Vanilla GKD & 84.24$\pm$0.27 & 28.80$\pm$0.80 & 97.27$\pm$0.33 & 71.20$\pm$0.80 \\
\method{} & 86.35$\pm$0.71 & 22.35$\pm$1.39 & 95.05$\pm$0.85 & 77.65$\pm$1.39 \\
\Supportunion{} & 90.08$\pm$0.13 & 11.08$\pm$1.16 & 91.23$\pm$0.94 & 88.92$\pm$1.16 \\
\bottomrule
\end{tabular}
\caption{Llama-3.1-8B APIGen-MT cross-family replication (percent). GKD rows
report mean$\pm$sample std over three seeds; Mixed SFT is a separately trained
single reference.}
\label{tab:robustness-main}
\end{table}

The Llama vanilla operating point is call-heavy and consistent across seeds.
Compared descriptively with the separately trained mixed-SFT reference,
vanilla records 5.37 points higher call recall, 18.55 points higher
over-calling, and 6.59 points lower decision accuracy. This reference does not
isolate a training effect because its objective and schedule differ. In the
matched GKD comparison, \method{} reduces over-calling by
6.45$\pm$0.64 points and call recall by 2.22$\pm$1.05 points; support union
changes them by 17.72$\pm$1.32 and 6.03$\pm$1.07 points. The two training
layers therefore reproduce distinct restraint--capability shifts. The
support-level run remains below mixed SFT on
When2Call, and its BFCL mean has substantial seed variation. Full Llama
reference and held-out results are in Appendix~\ref{app:llama-replication}.

%% file: sections/07_related_limitations.tex
\section{Related Work and Limitations}

\paragraph{On-policy and selective distillation.}
GKD supervises student rollouts with token-level teacher divergences
\citep{agarwal2024gkd}. Related work studies reverse-KL distillation,
self-distillation, entropy-aware objectives, global normalization, trajectory
filtering, and token reweighting
\citep{xu2024minillm,ko2025distillm2,zhao2026opsd,jin2026eopd,hao2026gndpo,li2026fireopd}.
TIP prioritizes response positions using student entropy and teacher--student
disagreement \citep{tip2026}. Teachability-Aware OPD also forms a
teacher/student top-$K$ union to score sequence-position compatibility
\citep{wang2026teachability}. We instead hold supervised positions and routing
fixed, then use the union as the actual within-position support of the
truncated divergence.

\paragraph{Sparse and truncated logit support.}
Sparse Logit Sampling shows that caching only teacher top-$K$ probabilities
gives a biased distribution estimate and uses importance-sampled tail logits
to preserve the full gradient in expectation \citep{anshumann2025sparse}.
Tail-Aware Distillation instead separates teacher modes from the full-vocabulary
tail and amplifies the latter's aggregate contribution
\citep{dasgupta2026tail}. Analyses of top-$K$-censored APIs characterize the
remaining distributional ambiguity, while communication-aware distillation
uses adaptive $K$ to reduce high-dimensional logit transfer
\citep{nie2026identified,zhang2025communication}. These works establish that
sparse logits and low-mass tails require care. Our focus is complementary:
which omitted coordinate corrects a student's discrete behavior, how routed
teachers make the omission directional, and whether restoring that coordinate
changes complete generations.

\paragraph{Multi-teacher routing and conflict.}
Multi-teacher OPD studies teacher routing, specialization, and capability
integration \citep{opdsurvey2026,ma2026mopd,wang2026madopd}. EWAD routes
supervision using reliability and agreement \citep{sumit2026ewad};
Counteraction-Aware MOPD separates conflicting recovery and preservation
updates \citep{chen2026camopd}. We hold routing fixed and show that the routed
loss itself can be directionally incomplete: the two teachers need not expose
the same decision token in their truncated supports.

\paragraph{Loss shaping and deployment calibration.}
Multi-objective training balances losses or gradients
\citep{kendall2018multitask,chen2018gradnorm}, while distillation methods
reshape token contributions \citep{jung2025todi,xie2026adakd,tip2026}.
Our diagnostic restoration interventions reinstate an omitted coordinate rather than
reweighting existing ones. The optimization study then treats student-aware
support construction, Hard Clip / Global Reweight / \method{} loss shaping,
and entry bias as parallel support-, loss-, and decoding-level strategies.

\paragraph{Tool-use behavior drift.}
Recent work reports tool overuse, collapse, and repeated calls in RL or
agentic search
\citep{hao2026toolrlcollapse,zeng2026tooloveruse,xue2025simpletir,qian2025smart,jin2025searchr1,deng2025grpoCollapseSearchR1}.
We identify a distinct route to related behavior inside supervised
multi-teacher OPD: a low-probability structural token is removed from one
teacher's support, and the resulting one-sided update moves the shared
call/respond boundary.

\paragraph{Limitations.}
The matched causal chain---forced replay, exact restoration, and temporal-scope
training---is scoped to Qwen3.5 and the implemented
teacher-top-32 objective. The bilateral frozen audit establishes the sign of
the Qwen routed entry updates, but it is an endpoint logit-space measurement,
not a full-vocabulary training experiment or proof that one coordinate fully
mediates the learned behavior. In particular, tool top-32 preserves the
reinforcing direction while substantially attenuating its full-vocabulary
magnitude. Llama-3.1 independently reproduces a call-heavy vanilla operating
point, response-side support omission, and matched support-aware correction
under a native JSON protocol, but its frozen support audit is not a second
matched restoration study. Both families use the same tool-call/response task
construction. Exact all-position restoration is deliberately broad and
suppresses required as well as unnecessary calls. The placebo matches scope,
support count, and teacher probability, but not student probability or
effective gradient strength. The gradient-reroute control matches the forward
loss and selected output-logit gradient magnitude, but not the output row,
hidden-state contribution, or full parameter gradients; it also degrades BFCL
single-turn and When2Call relative to vanilla. The optimization layers
also require different access: support uses teacher probes, loss shaping
cannot restore omitted coordinates, and entry bias needs decoder control.
Finally, BFCL multi-turn
is a controlled harness whose exact metric does not judge final-answer
semantics, and its supported endpoint summary excludes a locally unimplemented
category.

\paragraph{Artifact availability.}
Training code, analysis scripts, evaluation harnesses, aggregate metrics, and
plotting scripts are available in our
\href{https://github.com/shen-jiabin/decision-support-opd}{public repository}.

%% file: sections/08_conclusion.tex
\section{Conclusion}

Teacher top-$K$ support can retain nearly all probability mass while losing
a correction needed for the student's behavior. In Qwen's routed tool-use
setting, response top-32 usually omits the entry coordinate, while tool top-32
preserves its reinforcing sign but attenuates its full-vocabulary magnitude.
Frozen audits, forced-entry replay, and matched restoration connect this
asymmetry to behavior. First-position restoration relocates calls;
all-position restoration reduces over-calling from 14.2\% to 3.7\% while also
suppressing required calls. Teacher-probability matching and selected
output-logit-gradient rerouting do not reproduce that effect, supporting
coordinate omission as a causal contributor within the tested setting.

The intervention study answers the next question: which changes are useful
under a given error preference and system interface? Support union recovers
student-relevant coordinates at additional query and training cost. Loss
shaping changes retained signals through retraining, and entry bias offers a
validation-tuned decoding adjustment. Their comparison separates modest
changes in decision separation from larger operating-point shifts. Error
costs change their relative preference, and multi-turn results show why fewer
loops must be assessed together with required-call coverage and dialogue
success.

Llama-3.1 reproduces the directional support asymmetry and matched
intervention pattern under native JSON. Together, these results motivate
checking both what behavioral corrections compressed supervision preserves
and what each intervention costs to recover useful behavior.

%% file: sections/09_appendix.tex
\section{Additional Experimental Details}

\subsection{Training data and supervision}

The main GKD experiments use two filtered training splits derived from APIGen-MT: 15,419 tool-call examples and 15,245 response examples. We first split 5,000 APIGen-MT conversations at the conversation level with split seed 42: 3,500 conversations for training, 500 for validation, and 1,000 for test. The conversation IDs are disjoint across splits. All GKD variants use the same post-filter files. The validation split is approximately balanced, with 2,188 tool-call examples and 2,172 response examples. The APIGen-MT-derived decision test is extracted only from the held-out test conversations, with four balanced decision buckets of 1,000 examples each, for 4,000 total test decision points. Each GKD training example is routed to one teacher by its teacher tag. The tool-call teacher supervises examples whose final target is a structured tool call, and the response teacher supervises examples whose final target is a natural-language answer.

Only the final assistant turn is supervised; earlier turns provide context. Because our ms-swift/Megatron backend consumes this target as text, we pre-render tool calls in the model template's XML-style form using tags such as \tooltok{}, \texttt{<function=...>}, and \texttt{<parameter=...>}. This backend-specific step prevents unrendered objects from corrupting the learned schema.

\subsection{Tool-call target rendering}
\label{app:tool-call-rendering}

For structured tool-call examples, the final supervised target is rendered as text before training in our backend. A representative target is:
{\small
\setlength{\topsep}{2pt}
\setlength{\partopsep}{0pt}
\begin{verbatim}
<tool_call>
<function=get_flight_cost>
<parameter=travel_to>
LAX
</parameter>
</function>
</tool_call>
\end{verbatim}
}
This appendix example is illustrative; the experiments use the rendered targets from the APIGen-MT-derived training split. Other frameworks may serialize the same structured call differently.

\subsection{Main training configuration}

Table~\ref{tab:appendix-training-config} lists the configuration shared by the
four loss-comparison variants and the support-construction interventions. All
runs use the same data, teachers, training length, and batch settings within
each seed. The loss-comparison variants differ only in their modifier; exact
restoration and its matched controls append one matched coordinate, while the
\supportunion{} probes student-selected coordinates as described in
Sections~\ref{sec:support-injection-training} and
\ref{sec:intervention-layers}.

\begin{table}[!htbp]
\centering
\small
\begin{tabular}{ll}
\toprule
Item & Value \\
\midrule
Initialization checkpoint & Qwen/Qwen3.5-9B \\
Training backend & ms-swift Megatron GKD backend \\
Precision & bfloat16 \\
Epochs & 1 \\
Micro batch size & 4 \\
Global batch size & 96 \\
Maximum input length & 14,000 tokens \\
Maximum completion length & 512 tokens \\
Learning rate & $5\times 10^{-6}$ \\
Warmup fraction & 0.1 \\
Tensor parallel size & 8 \\
GKD rollout mixture $\lambda$ & 0.8 \\
GKD $\beta$ & 0.5 \\
Distillation / rollout temperature & 0.9 \\
Rollout top-$p$ / top-$k$ & 1.0 / disabled \\
Sequence KD & false \\
Supervised format anchor & \texttt{sft\_alpha}=0.3 \\
Teacher logit top-$K$ & 32 \\
Loss scale & \texttt{last\_response+ignore\_empty\_think} \\
Loss-comparison checkpoints & Every 50 steps \\
Support-intervention checkpoints & Final epoch \\
\bottomrule
\end{tabular}
\caption{Shared GKD training configuration.}
\label{tab:appendix-training-config}
\end{table}

All GKD variants share the supervised anchor for structured-output stability. Student rollouts are sampled with probability $\lambda=0.8$; the SFT anchor applies only to dataset-source trajectories, as defined in Section~\ref{sec:setup}.

\paragraph{Teacher and SFT reference training.}
The 9B tool-call teacher, response teacher, and Mixed SFT reference are full fine-tunes from the Qwen3.5-9B checkpoint. They use the same ms-swift SFT backend, bfloat16 precision, DeepSpeed ZeRO-3, Qwen3.5 template, maximum length 10,000, learning rate $5\times10^{-6}$, cosine schedule with 0.1 warmup fraction, micro batch size 4, gradient accumulation 2, and two epochs. The tool-call teacher is trained only on the tool-call SFT split, the response teacher only on the response SFT split, and Mixed SFT on the mixed SFT split. The 4B Mixed SFT reference uses the same recipe with Qwen3.5-4B. We select the epoch-2 checkpoints used throughout the evaluation.

\paragraph{Loss-modifier hyperparameters.}
Before multi-seed test evaluation, we fix \method{} at $k=3$, Hard Clip at $c=0.5$, and Global Reweight at $\alpha=0.3$, $z_{\max}=3.0$, $w_{\min}=0.25$, and $w_{\max}=2.0$. The method comparison uses these fixed settings.

\FloatBarrier
\section{Metric Definitions}

\subsection{Evaluation metrics}

For the APIGen-MT-derived decision set, \emph{decision accuracy} is the
fraction of examples for which the model makes the correct tool-use mode-entry
decision. We remove thinking blocks and apply a family-specific parser before
normalizing detected calls to one canonical representation. The Qwen parser
detects its XML \tooltok{} block; the Llama parser detects a native JSON object
with \texttt{name} and \texttt{parameters} fields. \emph{Over-calling} is the
fraction of should-respond examples for which the corresponding parser detects
a call. \emph{Call recall} is detection recall on should-call examples, and
\emph{respond recall} is recall on should-respond examples. These decision
metrics do not require schema-valid or executable arguments; schema validity
is instead tested by the separate BFCL call-quality evaluation. The APIGen-MT-derived decision test used in this paper
is balanced, so decision accuracy is the average of call recall and respond
recall, while over-calling equals one minus respond recall. We retain all
columns to make the call/response operating-point trade-off explicit. In this
paper, boundary calibration means operating-point calibration between
tool-call and direct-response decisions; it is not probability calibration
measured by ECE, Brier score, or reliability diagrams.

\paragraph{Decoding settings.}
Main single-turn APIGen-MT, BFCL, and When2Call inference uses a vLLM-backed OpenAI-compatible server with thinking disabled, greedy decoding (\texttt{temperature}=0, \texttt{top\_p}=1), and dataset-specific generation lengths: 512 new tokens for APIGen-MT, 256 for BFCL, 16 for When2Call multiple choice, and 512 for When2Call judge-style outputs. The BFCL multi-turn diagnostic also uses greedy decoding, a maximum of five tool-call steps per user turn, and a fixed simulated tool-observation harness. The strict E1 decoding evaluation in Appendix~\ref{app:e1-boundary-bias} instead uses a local Transformers decoder so that a first-token-only bias can be applied; that bias is applied at the first generated assistant token of every assistant generation step.

For BFCL, we use local BFCL v4-style processed data from seven single-turn subsets: simple-python, multiple, parallel, live-simple, live-multiple, irrelevance, and live-irrelevance. The processed files contain 400, 200, 200, 258, 1053, 240, and 884 examples, respectively. Our project-local scorer checks exact AST-style matches for tool-call subsets, allowing any acceptable BFCL ground-truth value, and counts irrelevance examples as correct when no tool call is parsed. These are local BFCL-v4-style scores rather than official leaderboard submissions.

For When2Call, we use a fixed 1,000-example subset sampled with seed 42 from
the 3,652-example public processed MCQ snapshot. The evaluated subset contains
329 tool-call, 289 request-for-information, and 382 cannot-answer targets. The
four choices map A to direct response, B to tool call, C to request for
information, and D to cannot answer; the scorer extracts the final standalone
A--D answer letter. The raw 300-example judge-style file is preprocessed with
a 512-token generation length for completeness, but the reported tables use
only the MCQ protocol.

\subsection{Training process metrics}

The sample-level diagnostic records contain both raw and adjusted statistics. Raw statistics are computed before any loss modifier. Adjusted statistics are computed after Soft Clamp, Hard Clip, or Global Reweighting. In vanilla GKD, raw and adjusted values are identical. We compute every reported tag-level quantity from records that preserve the pre-packing sample boundary; packed-row aggregate tag fields are not used.

The main aggregate diagnostic is the raw per-token top-$K$ JSD ratio:
\begin{equation}
    R_{\mathrm{JSD}} =
    \frac{\sum_{i\in \mathrm{tool}} d_i / N_{\mathrm{tool}}}
         {\sum_{i\in \mathrm{resp}} d_i / N_{\mathrm{resp}}},
\end{equation}
where $d_i$ is the token-level top-$K$ divergence and $N$ is the number of supervised loss tokens. In teacher-API runs, teacher log probabilities are returned for the top 32 teacher tokens; student logits are gathered on those token IDs, and both distributions are renormalized on that truncated support before computing JSD. We also use a squared-divergence proxy ratio:
\begin{equation}
    R_{\mathrm{grad}} =
    \frac{\sum_{i\in \mathrm{tool}} d_i^2 / N_{\mathrm{tool}}}
         {\sum_{i\in \mathrm{resp}} d_i^2 / N_{\mathrm{resp}}}.
\end{equation}
This proxy is not a parameter-gradient norm; it is a scale-sensitive diagnostic for whether a small number of large token divergences dominate the divergence mass. These ratios remove the first-order effect of response length and batch composition. Values below one mean that response examples have larger average full-sequence divergence or squared-divergence proxy.

Decision-boundary metrics inspect the first supervised token. We log the student's probability of \tooltok{}, its log-odds margin over non-tool tokens and common response starters, and whether it is top-1. These metrics track entry pressure; full-generation evaluation captures later behavior.

\FloatBarrier
\subsection{Aggregate sanity diagnostics}
\label{app:aggregate-sanity}

Table~\ref{tab:aggregate-sanity} gives packed-safe sampled exposure and divergence diagnostics from representative runs. T/R denotes tool-call divided by response. Each method contributes 31 paired diagnostic steps, with one tool-call and one response sample retained every ten steps. Intervals use paired bootstrap resampling over steps.

\begin{table}[!htbp]
\centering
\scriptsize
\setlength{\tabcolsep}{3pt}
\begin{tabular}{lccc}
\toprule
Method & Sampled token T/R & Raw per-token JSD T/R & Raw squared-div. T/R \\
\midrule
Vanilla GKD & 0.813 [0.650, 1.013] & 1.031 [0.734, 1.434] & 1.339 [0.810, 2.168] \\
Hard Clip & 0.811 [0.637, 1.035] & 1.014 [0.692, 1.487] & 1.301 [0.742, 2.215] \\
Global Reweight & 0.807 [0.644, 1.008] & 0.950 [0.679, 1.296] & 1.187 [0.741, 1.848] \\
\method{} & 0.934 [0.738, 1.178] & 0.983 [0.675, 1.404] & 1.214 [0.706, 1.953] \\
\bottomrule
\end{tabular}
\caption{Packed-safe sampled sanity checks from representative diagnostic runs. Brackets are paired 95\% bootstrap intervals. Every interval includes one, so these sparse full-sequence diagnostics do not establish a stable direction of teacher dominance.}
\label{tab:aggregate-sanity}
\end{table}

\FloatBarrier
\subsection{Token-level concentration diagnostics}
\label{app:token-concentration}

Table~\ref{tab:appendix-token-concentration} gives the multi-seed numerical values behind the concentration discussion in Section~\ref{sec:diagnosis}. The concentration columns report the average share of per-sample JSD carried by the largest 1\%, 5\%, and 10\% of supervised tokens. The shrinkage column reports the mean adjusted squared-divergence proxy divided by the raw squared-divergence proxy on diagnostic samples. The threshold-event column reports the fraction of diagnostic tokens that explicitly exceed a hard or soft clamp threshold; it is not defined for vanilla GKD or Global Reweight, because Global Reweight modifies token weights broadly rather than producing explicit clamp events.

\begin{table}[!htbp]
\centering
\small
\begin{tabular}{lccccc}
\toprule
Method & Top 1\% JSD & Top 5\% JSD & Top 10\% JSD & Proxy shrinkage & Threshold events \\
\midrule
Vanilla GKD & 0.412$\pm$0.004 & 0.768$\pm$0.029 & 0.910$\pm$0.023 & 1.000$\pm$0.000 & N/A \\
Hard Clip & 0.389$\pm$0.003 & 0.763$\pm$0.026 & 0.907$\pm$0.022 & 0.972$\pm$0.002 & 0.003$\pm$0.001 \\
Global Reweight & 0.327$\pm$0.024 & 0.672$\pm$0.015 & 0.859$\pm$0.018 & 0.462$\pm$0.040 & N/A \\
\method{} & \textbf{0.154$\pm$0.032} & \textbf{0.507$\pm$0.021} & \textbf{0.794$\pm$0.031} & 0.283$\pm$0.047 & 0.074$\pm$0.005 \\
\bottomrule
\end{tabular}
\caption{Multi-seed diagnostic token-level signal concentration and compression. Values are mean$\pm$std over seeds 42, 44, and 60. Lower concentration indicates that fewer extreme tokens dominate the token-level JSD mass.}
\label{tab:appendix-token-concentration}
\end{table}

\FloatBarrier
\subsection{Mode-mismatch replay and decision-support audit}
\label{app:mode-mismatch-support}

We use the APIGen-MT validation split, with 500 tool-tagged and 500
response-tagged prompts. Each of the three vanilla checkpoint-319 students
samples with thinking disabled, temperature 0.9, top-$p=1$, and at most 512
new tokens. We replay the resulting 3,000 trajectories under their routed
teachers with the training objective: beta 0.5 JSD, teacher-top-32 support,
and support renormalization. Confidence intervals use 10,000 prompt bootstrap
replicates.

\begin{table}[!htbp]
\centering
\small
\begin{tabular}{lrrrr}
\toprule
Student seed & Tool mismatch & Tool JSD ratio & Response mismatch & Response JSD ratio \\
\midrule
42 & 5.8\% & 50.63 [32.22, 89.83] & 11.6\% & 2.40 [1.82, 3.09] \\
44 & 4.6\% & 57.15 [38.41, 94.72] & 12.8\% & 1.76 [1.23, 2.39] \\
60 & 5.6\% & 64.39 [36.94, 130.31] & 10.4\% & 1.96 [1.49, 2.57] \\
Pooled & 5.3\% & 55.45 [41.74, 76.44] & 11.6\% & 2.03 [1.71, 2.39] \\
\bottomrule
\end{tabular}
\caption{Natural mode-mismatch replay. A tool mismatch is a response-mode rollout under the tool teacher; a response mismatch is a tool-mode rollout under the response teacher. JSD ratios divide mismatched by aligned per-token JSD. Brackets are 95\% bootstrap intervals.}
\label{tab:natural-mode-mismatch}
\end{table}

\FloatBarrier
Natural mismatch may select harder prompts. We therefore sample 200 prompts
per teacher tag for seed 42 and create paired continuations: one forces
\tooltok{} at the first generated position, while the other masks that token
and samples a response entry. The student generates all later tokens normally.
Under the tool teacher, mismatched JSD is 0.029140 per token versus 0.001090
when aligned, a ratio of 26.73 [14.87, 67.91]; 99.5\% of prompts have higher
mismatched JSD. Under the response teacher, the corresponding values are
0.013040 and 0.004886, a ratio of 2.67 [2.29, 3.14]; 87.5\% of prompts are
higher under mismatch.

\begin{table}[!htbp]
\centering
\small
\begin{tabular}{lrrrr}
\toprule
Teacher $\rightarrow$ student mode & First 1 & First 4 & Tokens 5--10 & After 10 \\
\midrule
Tool $\rightarrow$ tool & 0.00266 & 0.00079 & 0.00193 & 0.00064 \\
Tool $\rightarrow$ response & 0.00279 & 0.06407 & 0.07706 & 0.03220 \\
Response $\rightarrow$ response & 0.01407 & 0.00817 & 0.00672 & 0.00568 \\
Response $\rightarrow$ tool & 0.01395 & 0.10543 & 0.01758 & 0.01055 \\
\bottomrule
\end{tabular}
\caption{Per-token JSD by trajectory window in the forced-entry pairs. The first-position distributions are shared before the intervention. The mismatch appears after mode entry and extends beyond the first few tokens.}
\label{tab:forced-entry-windows}
\end{table}

\FloatBarrier
For the Qwen routed support audit, we recompute the first supervised position
with local full-vocabulary teacher and matched-student logits, one prompt at a
time. This avoids padding-dependent numerical changes and preserves 500 fixed
validation prompts per route for each seed. Teacher coverage, mass, and rank
use the 500 unique teacher distributions per route; student-dependent gradient
and union quantities use all 1,500 prompt--student pairs per route. All descent
quantities use $\tau=0.9$; reported teacher mass uses the raw pre-temperature
distribution. Mean response-teacher full-vocabulary $P(\tooltok{})$ is
$6.14\times10^{-8}$. Its strict competition rank is one plus the number of
strictly larger logits; it has median 236.5, with
90th, 95th, and 99th percentiles of 614.1, 781.9, and 1086.0; ranks span
25--1243. Although top-32 retains 0.999900 of teacher probability mass, it
contains \tooltok{} for only 0.4\% of prompts.

\begin{table}[!htbp]
\centering
\small
\begin{tabular}{lrrrr}
\toprule
Student seed & $K=32$ coverage & $K=32$ descent & $K=32$ + tool & Full vocabulary \\
\midrule
42 & 0.4\% & $-0.000001$ & $-0.037799$ & $-0.037749$ \\
44 & 0.4\% & $-0.000003$ & $-0.044787$ & $-0.044739$ \\
60 & 0.4\% & $-0.000003$ & $-0.039682$ & $-0.039636$ \\
Pooled & 0.4\% & $-0.000002$ & $-0.040756$ & $-0.040708$ \\
\bottomrule
\end{tabular}
\caption{Frozen response-side tool-logit descent, $-\partial d/\partial z_{\mathrm{tool}}$. Adding the omitted decision token to top-32 restores the full-vocabulary direction in every seed.}
\label{tab:support-injection-seeds}
\end{table}

Coverage is computed from the actual top-$K$ set. At a tied cutoff it can
differ from the strict competition-rank indicator because the latter assigns
all tied logits the same rank.

The teacher--student top-32 union contains \tooltok{} in all 1,500 matched
response-side pairs, averages 38.46 coordinates, retains 0.999911 raw teacher
mass, and gives descent $-0.040748$, close to full-vocabulary $-0.040708$.
The complementary tool route reverses the sign: the tool teacher ranks
\tooltok{} first on all 500 unique prompts, and top-32 descent is positive in
all 1,500 matched pairs. Mean tool-route descent grows from $+0.001167$ at
$K=32$ to $+0.003882$, $+0.006175$, and $+0.008982$ at
$K=64,128,256$, while full-vocabulary descent is $+0.020174$. Tool-side
teacher--student top-32 union averages 58.16 coordinates and reaches
$+0.020065$. Thus tool top-32 preserves the reinforcing direction but
substantially attenuates its full-vocabulary magnitude despite near-unit raw
teacher mass.

The intervention adapts to the student's current boundary error. Among 1,326
response-mode entries, mean student $P(\tooltok{})$ is 0.093 and injected
descent is $-0.0290$. Among 174 erroneous tool-mode entries, these values rise
to 0.702 and $-0.1303$. The injected-minus-full descent difference is
$-0.000048$ [${-0.000054}$, ${-0.000043}$], small relative to the full mean
magnitude. By contrast, $K=128$ minus full is 0.031549 [0.029098, 0.034049].
This audit isolates a local logit-space property of the objective; it does not
measure how shared parameters integrate that signal during training.

\subsection{Matched support-scope training}
\label{app:support-injection-training}

We train first-position and all-position support students for vanilla seeds
42, 44, and 60. All runs share initialization, teacher pair, routed data,
anchored GKD settings, 319-step budget, and evaluation inputs. On
response-tagged examples, the intervention appends \tooltok{} with its exact
teacher logit whenever top-32 omits it. The first-position variant changes
only the first effective supervised position; the all-position variant applies
the same rule at every supervised response position. Tool-tagged examples are
unchanged.

\paragraph{First-position restoration.}
The first-position intervention supplies a direct test of the entry gradient.
Table~\ref{tab:support-injection-training} reports its matched change from
vanilla.

\begin{table}[!htbp]
\centering
\scriptsize
\setlength{\tabcolsep}{3pt}
\begin{tabular}{lrrrrrr}
\toprule
Seed & $\Delta$E1 resp. entry & $\Delta$E1 call entry & $\Delta$APIGen over-call & $\Delta$Calls/turn & $\Delta$Loop@3 & $\Delta$Dialogue exact \\
\midrule
42 & $-10.80$ & $-11.85$ & $-0.40$ & $-0.101$ & $-1.98$ & $-0.63$ \\
44 & $-12.65$ & $-11.45$ & $-4.15$ & $-0.561$ & $-10.27$ & $-2.25$ \\
60 & $-9.75$ & $-10.15$ & $+2.40$ & $-0.172$ & $-2.49$ & $-0.75$ \\
Mean & $-11.07$ & $-11.15$ & $-0.72$ & $-0.278$ & $-4.91$ & $-1.21$ \\
\bottomrule
\end{tabular}
\caption{First-position support minus vanilla GKD. Entries are percentage
points except calls/turn. E1 measures first-token entry; APIGen measures
whether the complete generation contains \tooltok{}; exact success uses the
joint task-level definition in Appendix~\ref{app:bfcl-multiturn}.}
\label{tab:support-injection-training}
\end{table}

The first-token E1 shift is large and consistent but not class-selective. Mean response
entry error falls from 14.47\% to 3.40\%, while mean call entry falls from
91.72\% to 80.57\%. First-token accuracy changes by only $-0.04$ points and
AUC changes from 0.9692 to 0.9712. The intervention moves the operating point
instead of improving threshold-free separation.

Call-position counts explain the weaker full-generation result. Across 6,000
should-respond outputs, first-position calls fall from 805 (13.42\%) to 184
(3.07\%), while delayed calls rise from 47 (0.78\%) to 625 (10.42\%). The
no-call count changes only from 5,148 (85.80\%) to 5,191 (86.52\%). On 6,000
should-call outputs, first-position calls fall from 90.52\% to 77.68\%,
delayed calls rise from 0.93\% to 13.78\%, and the no-call rate remains 8.5\%.
Thus support injection often postpones entry rather than changing the final
mode.

The BFCL multi-turn reductions are consistent at the task level. The paired
95\% bootstrap intervals for calls/turn are $[-0.157,-0.044]$,
$[-0.632,-0.491]$, and $[-0.212,-0.130]$ for seeds 42, 44, and 60. The
corresponding Loop@3 intervals are $[-3.40,-0.55]$, $[-11.96,-8.64]$, and
$[-3.55,-1.43]$ percentage points. Broader capability metrics expose the
cost: BFCL single-turn tool-call quality falls from 79.41$\pm$1.54\% to
75.72$\pm$2.74\%, and BFCL single-turn overall falls by 1.69 points on average. When2Call
improves by 2.63 points, consistent with a more conservative call policy.
Overall dialogue exact success changes by $-0.63$, $-2.25$, and $-0.75$ points; its
paired intervals are $[-1.88,0.63]$, $[-3.75,-0.75]$, and $[-2.13,0.63]$.
Thus the loop reduction is consistent, whereas an endpoint-completion gain is
absent and the completion decrease is resolved only for seed 44.

\paragraph{All-position restoration.}
First-position restoration leaves later response positions outside the
support intervention. All-position restoration tests whether this temporal
scope explains the call migration. Table~\ref{tab:support-scope-full} gives the
complete-generation and multi-turn aggregates.

\begin{table}[!htbp]
\centering
\scriptsize
\setlength{\tabcolsep}{3pt}
\begin{tabular}{lrrr}
\toprule
Metric & Vanilla & First-position & All-position \\
\midrule
E1 response entry error & 14.47$\pm$1.53 & 3.40$\pm$0.26 & 3.63$\pm$0.24 \\
E1 tool-call entry recall & 91.72$\pm$1.22 & 80.57$\pm$0.68 & 81.30$\pm$0.51 \\
E1 boundary AUC & 0.9692$\pm$0.0023 & 0.9712$\pm$0.0014 & 0.9725$\pm$0.0010 \\
APIGen decision accuracy & 88.63$\pm$0.28 & 88.99$\pm$0.79 & 87.67$\pm$0.94 \\
APIGen over-calling & 14.20$\pm$2.08 & 13.48$\pm$1.27 & 3.73$\pm$0.51 \\
APIGen call recall & 91.45$\pm$1.68 & 91.47$\pm$0.50 & 79.07$\pm$1.73 \\
BFCL single-turn overall & 78.98$\pm$1.72 & 77.29$\pm$1.66 & 78.10$\pm$1.56 \\
BFCL tool-call quality & 79.41$\pm$1.54 & 75.72$\pm$2.74 & 73.41$\pm$2.66 \\
BFCL irrelevance refusal & 78.17$\pm$2.91 & 80.25$\pm$3.35 & 86.92$\pm$3.09 \\
When2Call MCQ & 65.57$\pm$1.07 & 68.20$\pm$0.72 & 67.73$\pm$1.27 \\
Calls/turn & 1.515$\pm$0.122 & 1.237$\pm$0.154 & 0.975$\pm$0.045 \\
Loop@3 & 15.08$\pm$3.01 & 10.17$\pm$2.43 & 6.82$\pm$0.22 \\
Non-tool final & 87.31$\pm$7.24 & 90.99$\pm$0.31 & 93.82$\pm$0.08 \\
Overall dialogue exact success & 7.04$\pm$0.31 & 5.83$\pm$1.01 & 3.38$\pm$0.50 \\
\bottomrule
\end{tabular}
\caption{Three-seed support-scope comparison. Values are percentages except
calls/turn and report mean$\pm$sample std. BFCL single-turn overall combines tool-call and
irrelevance subsets, so the all-position row's higher refusal offsets its
lower tool-call quality.}
\label{tab:support-scope-full}
\end{table}

Call positions expose the mechanism directly. Table~\ref{tab:support-call-position}
removes the initial thinking wrapper and separates calls that begin the answer
from calls that appear after natural-language tokens.

\begin{table}[!htbp]
\centering
\small
\setlength{\tabcolsep}{4pt}
\begin{tabular}{llrrr}
\toprule
Target & Training & First-position call & Delayed call & No call \\
\midrule
Should respond & Vanilla & 13.42 & 0.78 & 85.80 \\
& First-position & 3.07 & 10.42 & 86.52 \\
& All-position & 3.08 & 0.65 & 96.27 \\
\midrule
Should call & Vanilla & 90.52 & 0.93 & 8.55 \\
& First-position & 77.68 & 13.78 & 8.53 \\
& All-position & 77.23 & 1.83 & 20.93 \\
\bottomrule
\end{tabular}
\caption{Tool-call position over 6,000 complete generations per target class,
pooled across three seeds (percent). First-position restoration relocates
calls; all-position restoration removes most delayed calls in both classes.}
\label{tab:support-call-position}
\end{table}

Relative to first-position support, all-position support reduces APIGen
over-calling by 10.10, 8.35, and 10.80 points across seeds 42, 44, and 60. The
same seeds lose 12.75, 10.70, and 13.75 points of APIGen call recall. Loop@3
falls by 5.84, 1.02, and 3.19 points. The intervention therefore converts the
previous call relocation into a consistent full-generation effect, but does so
by moving the shared operating point beyond the first-position treatment.

E1 confirms that temporal coverage, rather than a larger initial boundary
shift, separates the two interventions. All-position response entry error is
3.63$\pm$0.24\%, tool-call entry recall is 81.30$\pm$0.51\%, and AUC is
0.9725$\pm$0.0010---all close to the first-position values of
3.40$\pm$0.26\%, 80.57$\pm$0.68\%, and 0.9712$\pm$0.0014. The large
complete-generation gap therefore opens after the first token.

The task-level decomposition gives the same conclusion. Empty-ground-truth
no-call accuracy progresses from 37.70$\pm$0.51\% for vanilla to
47.41$\pm$6.52\% for first-position and 61.49$\pm$4.54\% for all-position
support. Observed required-turn call coverage moves in the opposite direction:
92.90$\pm$0.33\%, 83.15$\pm$5.61\%, and 73.24$\pm$4.62\%. All-position
overall dialogue exact success is 3.38$\pm$0.50\%, versus 7.04$\pm$0.31\%
for vanilla and 5.83$\pm$1.01\% for first-position support. The support
coordinate is behaviorally causal but not response-selective under shared
parameters.

\subsection{Matched controls and student-aware support}
\label{app:support-construction-controls}

The probability-matched placebo uses the same all-position response scope as
exact restoration, but appends a non-tool tail token whose teacher probability
is closest to that of \tooltok{}. The support-level strategy appends all
student-top-32 tokens missing from teacher top-32 and queries their teacher
logits. Its within-position support is
$U_i^{(K)}=\operatorname{TopK}(p_t)_i\cup\operatorname{TopK}(p_s)_i$, and
the same restricted JSD is evaluated on $U_i^{(K)}$.
Across 319 steps and three seeds, teacher/student overlap is
25.12$\pm$0.08 tokens, so the union support averages only
38.88$\pm$0.08 rather than 64.
The returned support therefore grows by 21.5\% relative to teacher top-32.
Across the matched training logs, average steady-state step time rises from
106.4 to 123.5 seconds (16.0\%), while peak-memory differences remain within
run-to-run variation.

\begin{table}[!htbp]
\centering
\scriptsize
\setlength{\tabcolsep}{1.8pt}
\begin{tabular}{@{}lrrrrr@{}}
\toprule
Metric & Vanilla & Prob. placebo & Exact & Grad. reroute & \Supportunion{} \\
\midrule
E1 response entry error & 14.47$\pm$1.53 & 13.37$\pm$1.10 & 3.63$\pm$0.24 & 9.25$\pm$3.43 & 8.72$\pm$0.74 \\
E1 tool-call entry recall & 91.72$\pm$1.22 & 91.30$\pm$1.00 & 81.30$\pm$0.51 & 85.83$\pm$2.92 & 90.55$\pm$1.00 \\
E1 boundary AUC & 0.9692$\pm$0.0023 & 0.9695$\pm$0.0036 & 0.9725$\pm$0.0010 & 0.9609$\pm$0.0082 & 0.9760$\pm$0.0006 \\
APIGen over-calling & 14.20$\pm$2.08 & 13.25$\pm$1.96 & 3.73$\pm$0.51 & 11.65$\pm$2.88 & 7.38$\pm$0.64 \\
APIGen call recall & 91.45$\pm$1.68 & 90.37$\pm$2.31 & 79.07$\pm$1.73 & 86.93$\pm$3.31 & 87.02$\pm$2.01 \\
BFCL single-turn overall & 78.98$\pm$1.72 & 78.95$\pm$0.44 & 78.10$\pm$1.56 & 76.76$\pm$1.04 & 81.53$\pm$0.93 \\
When2Call MCQ & 65.57$\pm$1.07 & 68.53$\pm$0.92 & 67.73$\pm$1.27 & 26.70$\pm$22.32 & 68.27$\pm$1.11 \\
Calls/turn & 1.515$\pm$0.122 & 1.346$\pm$0.021 & 0.975$\pm$0.045 & 1.464$\pm$0.481 & 1.118$\pm$0.057 \\
Loop@3 & 15.08$\pm$3.01 & 12.44$\pm$1.41 & 6.82$\pm$0.22 & 15.78$\pm$9.97 & 8.05$\pm$0.87 \\
Required-turn call & 92.90$\pm$0.33 & 86.64$\pm$4.06 & 73.24$\pm$4.62 & 83.43$\pm$7.84 & 81.22$\pm$2.38 \\
Overall dialogue exact success & 7.04$\pm$0.31 & 4.92$\pm$0.40 & 3.38$\pm$0.50 & 5.21$\pm$0.44 & 4.12$\pm$0.33 \\
\bottomrule
\end{tabular}
\caption{Three-seed diagnostic-control and support-construction comparison
(percent except AUC and calls/turn; mean$\pm$sample std). Gradient rerouting
matches the forward loss and selected output-logit-gradient magnitude but does
not reproduce exact-restoration behavior; student-aware union is the
support-level strategy.}
\label{tab:support-construction-controls}
\end{table}

The placebo controls support count, scope, and teacher-probability scale, but
not student probability or effective gradient strength. It is not inert: its
loop rate and overall dialogue exact success also move. Exact all-position
restoration is a strong diagnostic intervention, not a recommended objective.
A separate response-side tool-entry probability penalty leaves the original
teacher-top-32 support unchanged and adds
\begin{equation}
\mathcal{L}_{\mathrm{penalty}}
=-\frac{\eta}{|\mathcal{I}_{\mathrm{resp}}|}
\sum_{i\in\mathcal{I}_{\mathrm{resp}}}\log(1-q_i(\tooltok{})),
\qquad \eta=0.15,
\end{equation}
where $\mathcal{I}_{\mathrm{resp}}$ contains all supervised response positions
and $q_i$ is the full-vocabulary softmax of the student logits used by the
penalty. The implementation clamps $q_i$ at $1-10^{-6}$ for numerical
stability and returns zero if the batch has no eligible position. At this
single fixed weight, APIGen over-calling is 12.93$\pm$1.14\%, call recall is
90.40$\pm$2.05\%, and overall dialogue exact success is 4.96$\pm$0.07\%
across three matched seeds. This supplementary control is not a tuned penalty
comparison or a match to exact restoration's gradient.

As the student-aware support-level strategy, the \supportunion{} raises
mean APIGen decision accuracy, BFCL single-turn overall, and When2Call MCQ by
1.19, 2.55, and 2.70 matched points.
Its required-turn call coverage nevertheless falls by 11.69 points, explaining
the lower overall dialogue exact success.

\begin{table}[!htbp]
\centering
\scriptsize
\setlength{\tabcolsep}{3pt}
\begin{tabular}{lrrrr}
\toprule
Support union $-$ Vanilla & Seed 42 & Seed 44 & Seed 60 & Mean [95\% CI] \\
\midrule
APIGen decision accuracy & +1.87 & +0.98 & +0.72 & +1.19 [$-0.31$, +2.69] \\
APIGen over-calling & $-5.95$ & $-9.85$ & $-4.65$ & $-6.82$ [$-13.54$, $-0.09$] \\
E1 boundary AUC & +0.46 & +0.78 & +0.79 & +0.68 [+0.20, +1.15] \\
BFCL single-turn overall & +2.75 & +4.08 & +0.81 & +2.55 [$-1.54$, +6.63] \\
When2Call MCQ & +3.50 & +2.50 & +2.10 & +2.70 [+0.91, +4.49] \\
Required-turn call & $-12.41$ & $-13.29$ & $-9.36$ & $-11.69$ [$-16.81$, $-6.57$] \\
Overall dialogue exact success & $-3.38$ & $-3.13$ & $-2.25$ & $-2.92$ [$-4.38$, $-1.45$] \\
\bottomrule
\end{tabular}
\caption{Matched support-union deltas in percentage points. The interval is
a two-sided 95\% $t$ interval over the three training-seed deltas; with three
seeds it is descriptive rather than a high-powered significance test.}
\label{tab:topk-union-paired-deltas}
\end{table}

\FloatBarrier
\subsubsection{Output-logit-gradient-reroute control}
\label{app:gradient-reroute}

The synthetic output-logit-gradient-reroute control reuses the natural
placebo's all-position eligibility and legal non-tool candidate. Let $z_t$ and
$z_c$ be the student logits of the tool-entry token $t$ and selected candidate
$c$. The appended student value is the straight-through proxy
\begin{equation}
\begin{gathered}
\widetilde z_{t\rightarrow c}
=\operatorname{sg}(z_t)+z_c-\operatorname{sg}(z_c), \\
\widetilde z_{t\rightarrow c}=z_t\ \text{(forward)},\qquad
\frac{\partial\widetilde z_{t\rightarrow c}}{\partial z_t}=0,\qquad
\frac{\partial\widetilde z_{t\rightarrow c}}{\partial z_c}=1.
\end{gathered}
\end{equation}
Here $\operatorname{sg}$ stops gradients. The appended teacher value remains
the response-teacher log-probability of $t$, so the forward loss equals exact
restoration while its selected output-logit gradient is routed to $c$.

Of 1,500 frozen prompt--student pairs, 1,494 are eligible and all pass in every
seed. The remaining six are the same two prompts repeated across three student
seeds; response-teacher top-32 already contains $t$, so no restoration or
rerouting is applicable. On the eligible pairs, maximum forward-JSD and
coefficient errors are zero, and maximum L1/L2 logit-gradient norm relative
errors are $2.30\times10^{-7}$ and $1.19\times10^{-7}$.

The gradient-reroute-minus-exact APIGen over-calling differences are $+8.25$,
$+4.90$, and $+10.60$ points across seeds. This supports the identity of the
output coordinate as one contributor, but not full-parameter-gradient
isolation. For the selected coefficient $g$, the hidden-state contribution
changes from $gW_{\mathrm{tool}}$ to $gW_{\mathrm{candidate}}$. The
candidate/tool row-norm ratio has median 1.1525 and row cosine median
$-0.0172$; input and output embeddings are untied. The low, high-variance
When2Call result and lower BFCL single-turn score are substantive collateral
effects.

\clearpage
\section{Additional Benchmark Results}

\subsection{Additional APIGen-MT results}
\label{app:apigen-extra}

\begin{figure}[!htbp]
\centering
\includegraphics[width=0.82\linewidth]{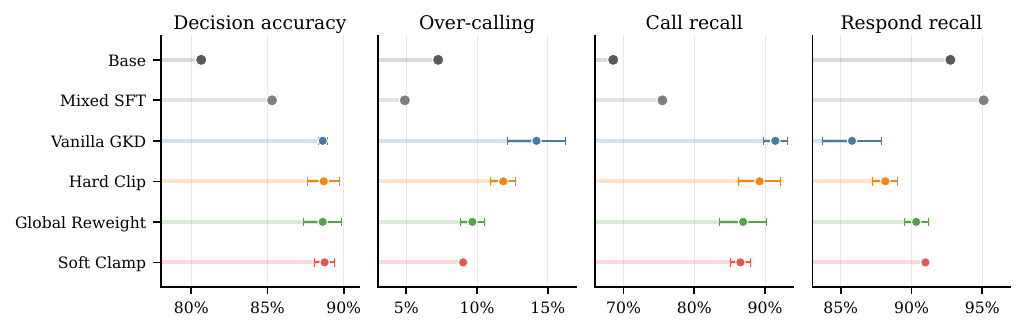}
\caption{APIGen-MT loss-shaping trade-off. Error bars mark standard deviation over three seeds. \method{} reduces over-calling relative to vanilla GKD while preserving decision accuracy, at the cost of lower call recall. Support construction is reported separately in Table~\ref{tab:support-construction-controls}.}
\label{fig:apigen}
\end{figure}

\begin{table}[!htbp]
\centering
\small
\setlength{\tabcolsep}{3pt}
\begin{tabularx}{\linewidth}{p{0.17\linewidth}X X X}
\toprule
Case & User state & Vanilla GKD & \method{} \\
\midrule
Airline baggage & User asks to add one checked bag and believes it is covered by free allowance. & Calls a baggage-update tool directly. & Explains the free-baggage allowance and asks for confirmation before updating. \\
Retail address & User asks to change Suite 716 to Suite 1000 at the same street address. & Calls an address-modification tool directly. & Summarizes the address change and asks the user to confirm. \\
Flight refund & User asks whether a completed trip can receive refund or credit due to membership status. & Starts another tool-call trajectory after stating the policy. & Directly explains that completed flights cannot be refunded or cancelled. \\
\bottomrule
\end{tabularx}
\caption{Qualitative APIGen-MT should-respond examples. Vanilla GKD enters a tool-call trajectory when the correct behavior is to answer, explain policy, or request confirmation. \method{} stays on the response side of the behavior boundary.}
\label{tab:case-study}
\end{table}

\FloatBarrier
\subsection{Decoding-level entry-bias evaluation}
\label{app:e1-boundary-bias}

We evaluate a non-positive scalar bias $b \in [-5,0]$ on vanilla GKD's
first-token \tooltok{} logit as a decoding-level strategy and test whether it
can reproduce \method{}'s boundary effect. With
thinking disabled, \emph{response tool-entry error} is the fraction of
should-respond examples whose top-1 first token is \tooltok{}; full-generation
APIGen over-calling remains a separate metric.

For each seed, we sweep $b$ from $-5$ to $0$ in steps of $0.05$ on validation
and select operating points matching either \method{}'s response entry error or
call recall. Ties use the smallest $|b|$; the bias is then frozen on test. E1
uses a common local Transformers scorer, whereas full-generation APIGen uses
vLLM, so the decoding comparison remains within its own protocol.

\begin{table}[!htbp]
\centering
\footnotesize
\begin{tabular}{lrrr}
\toprule
Setting & Response entry error & Tool-entry call recall & First-token acc. \\
\midrule
Vanilla GKD, no bias & 14.5$\pm$1.5 & 91.7$\pm$1.2 & 88.6$\pm$0.4 \\
\method{}, no inference bias & 11.1$\pm$0.3 & 90.4$\pm$0.2 & 89.7$\pm$0.1 \\
\Supportunion{}, no inference bias & 8.7$\pm$0.7 & 90.6$\pm$1.0 & 90.9$\pm$0.2 \\
Biased vanilla, val-matched response entry error & 11.2$\pm$0.6 & 89.2$\pm$1.2 & 89.0$\pm$0.6 \\
Biased vanilla, val-matched call recall & 11.6$\pm$1.1 & 89.5$\pm$0.6 & 88.9$\pm$0.5 \\
\bottomrule
\end{tabular}
\caption{Cross-layer first-token comparison on APIGen-MT test. Values are
percentages over three paired seeds, with sample standard deviations. Biases
are tuned on validation and frozen on test. The \supportunion{} changes training
support and is not bias-matched.}
\label{tab:e1-boundary-counterfactual}
\end{table}
\FloatBarrier

\begin{table}[!htbp]
\centering
\small
\begin{tabular}{lcc}
\toprule
Seed & Val-matched response-entry bias & Val-matched call-recall bias \\
\midrule
42 & -0.40 & -0.50 \\
44 & -0.65 & -0.40 \\
60 & -0.25 & -0.25 \\
\bottomrule
\end{tabular}
\caption{Validation-selected non-positive \tooltok{} entry biases used for
the decoding-level evaluation. Biases are selected on APIGen-MT validation
and frozen for APIGen-MT test and strict BFCL multi-turn evaluation.}
\label{tab:e1-selected-bias}
\end{table}
\FloatBarrier

Validation-tuned bias nearly matches \method{}'s response entry error (11.2\%
versus 11.1\%) but has lower call recall (89.2\% versus 90.4\%), lower accuracy
(89.0\% versus 89.7\%), and larger observed variability. The nearly unchanged
AUC reported in Section~\ref{sec:e1-main} supports the operating-point
interpretation for loss shaping. The support-level strategy lowers response
entry error to 8.7\% while retaining 90.6\% call-entry recall and increasing
AUC to 0.9760, consistent with student-aware support improving separation as
well as moving the threshold.

Section~\ref{sec:cost-main} reports the class-balanced risk definition,
validation-to-test selection for cost-aware bias, and the complete cost-sweep
figure. That analysis selects a separate bias for each seed and error-cost
ratio; the response-entry-matched and call-recall-matched biases in
Table~\ref{tab:e1-selected-bias} instead remain fixed for the strict-decoder
comparison below.

Vanilla GKD has 14.5$\pm$1.5\% first-token entry error and
14.3$\pm$2.2\% full-generation over-calling, close to the main-table
14.2$\pm$2.1\%. For \method{}, the values are 11.1$\pm$0.3\% and
9.0$\pm$0.2\%. The larger gap comes from cases that start with \tooltok{} but
do not retain the marker in the thinking-stripped output. Because E1 and full
generation use different backends, this disagreement is protocol-specific.

We also run a strict BFCL multi-turn decoding comparison with the same local
Transformers decoder, prompt rendering, parser, and bias hook for every row.
The main BFCL diagnostic uses vLLM, which lacks a first-token-only hook; backend,
batching, and template differences can change absolute values. We therefore
compare rows only within Table~\ref{tab:e1-strict-multiturn}.

\begin{table}[!htbp]
\centering
\footnotesize
\setlength{\tabcolsep}{3pt}
\begin{tabular}{lrrrrrr}
\toprule
Setting & Calls/turn & Loop@3 & Loop@5 & Repeat & Non-tool final & Invalid call \\
\midrule
Vanilla GKD, no bias & 1.734$\pm$0.057 & 20.5$\pm$1.5 & 15.1$\pm$1.6 & 23.7$\pm$4.3 & 78.8$\pm$7.0 & 2.3$\pm$2.2 \\
\method{}, no inference bias & 1.494$\pm$0.053 & 15.6$\pm$0.8 & 10.1$\pm$1.1 & 16.6$\pm$1.5 & 88.2$\pm$1.3 & 0.7$\pm$0.1 \\
Biased vanilla, val-matched response entry error & 1.492$\pm$0.089 & 15.7$\pm$1.0 & 10.3$\pm$1.2 & 17.6$\pm$0.6 & 85.8$\pm$1.5 & 1.6$\pm$1.2 \\
Biased vanilla, val-matched call recall & 1.525$\pm$0.062 & 16.3$\pm$1.3 & 11.0$\pm$1.3 & 18.8$\pm$4.0 & 84.7$\pm$4.8 & 1.8$\pm$1.5 \\
\bottomrule
\end{tabular}
\caption{Strict BFCL multi-turn decoding comparison with the same local
first-token-bias decoder for all rows. Values are three-seed mean$\pm$sample
std. The per-seed biases selected on APIGen-MT validation are reused here
without BFCL tuning. Inference-time scalar bias reproduces nearly all of the
mean reduction in tool-call frequency and loop rate under this strict decoder,
but \method{} retains higher non-tool-final rate and lower invalid-call rate.
Non-tool final is a termination metric, not answer correctness.}
\label{tab:e1-strict-multiturn}
\end{table}

Scalar bias nearly matches \method{} on call frequency and loop rate but leaves
worse mean non-tool-final and invalid-call rates. It is therefore a strong
decoding-level alternative; whether residual gaps persist under other
inference-time calibration protocols remains open.

\FloatBarrier
\subsection{BFCL and When2Call tables}
\label{app:bfcl-when2call}

\begin{table}[!htbp]
\centering
\small
\begin{tabular}{lrrr}
\toprule
Method & Overall & Tool Call Quality & Irrelevance Refusal \\
\midrule
Base & 82.2 & 83.5 & 79.8 \\
Mixed SFT & 82.9 & 82.0 & 84.4 \\
Vanilla GKD & 79.0$\pm$1.7 & 79.4$\pm$1.5 & 78.2$\pm$2.9 \\
Hard Clip & 79.9$\pm$0.6 & 78.1$\pm$1.2 & 83.2$\pm$1.2 \\
Global Reweight & 79.8$\pm$2.7 & 77.8$\pm$5.7 & 83.5$\pm$3.4 \\
\method{} & 80.8$\pm$0.1 & 79.7$\pm$1.3 & 82.7$\pm$2.4 \\
\Supportunion{} & 81.5$\pm$0.9 & 78.9$\pm$0.6 & 86.5$\pm$1.5 \\
\bottomrule
\end{tabular}
\caption{BFCL results. All values are percentages. GKD rows report
mean$\pm$std over three seeds; columns rank the strategies differently.}
\label{tab:bfcl}
\end{table}

\begin{table}[!htbp]
\centering
\small
\begin{tabular}{lrrrr}
\toprule
Method & MCQ Acc & tool\_call & request\_for\_info & cannot\_answer \\
\midrule
Base & 72.8 & 88.1 & 61.6 & 68.1 \\
Mixed SFT & 71.6 & 88.8 & 63.3 & 63.1 \\
Vanilla GKD & 65.6$\pm$1.1 & 88.0$\pm$1.4 & 58.9$\pm$5.8 & 51.2$\pm$4.1 \\
Hard Clip & 64.3$\pm$1.3 & 87.9$\pm$1.8 & 57.3$\pm$4.7 & 49.2$\pm$4.5 \\
Global Reweight & 67.1$\pm$2.2 & 87.3$\pm$2.5 & 61.5$\pm$6.9 & 53.8$\pm$3.4 \\
\method{} & 65.9$\pm$2.5 & 87.7$\pm$1.2 & 60.3$\pm$2.2 & 51.3$\pm$6.7 \\
\Supportunion{} & 68.3$\pm$1.1 & 87.6$\pm$2.0 & 59.8$\pm$6.2 & 58.0$\pm$4.6 \\
\bottomrule
\end{tabular}
\caption{When2Call results. All values are percentages. GKD rows report
mean$\pm$std over three seeds; category means rank the strategies
differently.}
\label{tab:when2call}
\end{table}

\FloatBarrier
\section{Secondary Qwen3.5-4B Student-Scale Check}
\label{app:qwen4b}

This secondary experiment replaces the Qwen3.5-9B student with Qwen3.5-4B
while retaining the fixed 9B teachers. It predates the support-level
interventions and therefore tests only whether the loss-shaping pattern
persists at a smaller student scale; it is not evidence for support-aware
correction.

\begin{table}[H]
\centering
\small
\begin{tabular}{lrrrr}
\toprule
Method & Dec. Acc & Over-call & Call Rec. & Resp. Rec. \\
\midrule
Mixed SFT & 88.5 & 7.5 & 84.5 & 92.5 \\
Vanilla GKD & 88.4$\pm$0.8 & 12.8$\pm$0.7 & 89.6$\pm$2.1 & 87.2$\pm$0.7 \\
Global Reweight & 87.8$\pm$0.8 & 9.5$\pm$0.8 & 85.1$\pm$1.9 & 90.5$\pm$0.8 \\
\method{} & 87.6$\pm$0.6 & 9.2$\pm$1.2 & 84.4$\pm$1.8 & 90.8$\pm$1.2 \\
\bottomrule
\end{tabular}
\caption{Qwen3.5-4B APIGen-MT decision results (percent). GKD rows report
mean$\pm$sample std over seeds 42, 44, and 60; Mixed SFT is a single
reference.}
\label{tab:qwen4b-secondary}
\end{table}

Global Reweight and \method{} reduce over-calling from 12.8$\pm$0.7\% to
9.5$\pm$0.8\% and 9.2$\pm$1.2\%. Their Loop@3 rates are 8.9$\pm$1.4\% and
9.2$\pm$2.6\%, versus 15.7$\pm$3.2\% for vanilla; neither raises overall
dialogue exact success. These results retain a scale check for loss shaping
without conflating it with the paper's support-level mechanism claim.

\FloatBarrier
\section{Llama-3.1 Cross-Family Replication}
\label{app:llama-replication}

The Llama replication uses Meta-Llama-3.1-8B-Instruct for the student and both
specialized teachers. Tool-call SFT renders native assistant JSON of the form
\verb|{"name": ..., "parameters": ...}|, while tool observations use
the model's native tool-response role. Mixed, response-only, and tool-only SFT
checkpoints train for two epochs. Mixed SFT is a separate supervised
reference; Vanilla GKD, \method{}, and the \supportunion{} run each initialize from
the original Instruct checkpoint and share one epoch, seeds 42/44/60, teacher
top-32, $\lambda=0.8$, $\alpha=0.3$, $\beta=0.5$, $\tau=0.9$, and a 512-token rollout limit. The
evaluation renderer and normalizer preserve this native protocol before
mapping calls to the common metric representation.

\begin{table}[H]
\centering
\scriptsize
\setlength{\tabcolsep}{2.5pt}
\begin{tabular}{lrrrrrr}
\toprule
Model & Dec. Acc & Over-call & Call Rec. & Resp. Rec. & When2Call & BFCL \\
\midrule
Base & 57.13 & 59.65 & 73.90 & 40.35 & 46.10 & 47.94 \\
Mixed SFT & 90.83 & 10.25 & 91.90 & 89.75 & 37.80 & 42.29 \\
Response SFT & 50.00 & 0.00 & 0.00 & 100.00 & 14.80 & 34.74 \\
Tool-call SFT & 50.00 & 100.00 & 100.00 & 0.00 & 33.80 & 45.29 \\
Vanilla GKD & 84.24$\pm$0.27 & 28.80$\pm$0.80 & 97.27$\pm$0.33 & 71.20$\pm$0.80 & 12.93$\pm$3.86 & 52.27$\pm$8.24 \\
\method{} & 86.35$\pm$0.71 & 22.35$\pm$1.39 & 95.05$\pm$0.85 & 77.65$\pm$1.39 & 12.77$\pm$3.93 & 48.75$\pm$8.26 \\
\Supportunion{} & 90.08$\pm$0.13 & 11.08$\pm$1.16 & 91.23$\pm$0.94 & 88.92$\pm$1.16 & 17.27$\pm$4.16 & 53.59$\pm$8.02 \\
\bottomrule
\end{tabular}
\caption{Llama-3.1-8B single-turn results (percent). GKD rows report
mean$\pm$sample std over three seeds; Base and SFT rows are single references.
The one-sided SFT teachers verify the intended routing extremes. BFCL varies
substantially across GKD seeds, and all GKD variants remain below mixed SFT on
When2Call.}
\label{tab:llama-full-results}
\end{table}

The frozen response-side support audit samples 500 unique response prompts and
evaluates the same prompts against vanilla students from all three seeds. A
complementary tool-side audit repeats this protocol on 500 unique tool prompts
with the tool teacher. The configured entry prefix \verb|{"name"| begins with
\llamaentry{} (ID 5018), which is the audited behavior coordinate. Teacher-only support metrics use the 500 unique
prompts per route. Support-union and gradient metrics pool 1,500 matched
teacher--student pairs per route.
All entry descents use $\tau=0.9$; reported teacher mass is the raw
pre-temperature probability mass. Ranks are one-indexed strict competition
ranks.

\begin{table}[H]
\centering
\small
\begin{tabular}{lrrr}
\toprule
Support & Entry coverage & Teacher mass & Entry-logit descent \\
\midrule
Response $K=32$ & 0.00\% & 0.999770 & 0.000000 \\
Response $K=64$ & 0.00\% & 0.999899 & 0.000000 \\
Response $K=128$ & 0.00\% & 0.999947 & 0.000000 \\
Response $K=256$ & 0.00\% & 0.999968 & 0.000000 \\
Response T$\cup$S top-32 & 97.93\% & 0.999786 & $-0.049229$ \\
Response full vocabulary & 100.00\% & 1.000000 & $-0.049180$ \\
\midrule
Tool $K=32$ & 100.00\% & 1.000000 & $+0.005174$ \\
Tool full vocabulary & 100.00\% & 1.000000 & $+0.009028$ \\
\bottomrule
\end{tabular}
\caption{Frozen Llama routed support audit. Entry descent is
$-\partial d/\partial z_{\mathrm{entry}}$; negative values lower the JSON-entry
logit and positive values raise it. Response-teacher top-32 retains 99.977\%
mass while omitting the decision coordinate on every response prompt;
student-aware union recovers nearly the full-vocabulary correction with 40.49
coordinates on average. Tool-teacher top-32 contains the coordinate on every
tool prompt and directly reinforces it.}
\label{tab:llama-support-audit}
\end{table}

The response teacher's exact entry rank spans 10,381--127,884, with median
105,952.5 and 90th percentile 122,199.9. Union entry coverage is 99.2\%,
97.6\%, and 97.0\% for vanilla student seeds 42, 44, and 60. These endpoint
students assign enough probability to the omitted coordinate for it to enter
their top-32, which is precisely the error teacher-only support cannot
directly correct. By contrast, the tool teacher ranks the entry token first on
all 500 tool prompts, with mean full-vocabulary probability 0.999999; its
top-32 entry descent is positive for all three matched students. The trained
support-union models' APIGen results in
Table~\ref{tab:robustness-main} provide the corresponding behavioral test;
the broader matched temporal-scope causal chain remains the Qwen primary
study.

\section{Loss Modifier Details}
\label{app:objective-details}

\subsection{Vanilla GKD}

Vanilla GKD uses the original token divergence $d_i$ without additional calibration. It is the reference point for measuring whether a loss modifier reduces behavior imbalance.

\subsection{Hard Clip}

Hard Clip applies
\begin{equation}
    d'_i = \min(d_i, c),
\end{equation}
with $c=0.5$ in the main experiment. This baseline tests whether simply truncating extreme token losses can reduce over-calling. It is a practical contrast rather than a mechanism-matched control: it uses a fixed threshold, removes marginal gradients above the threshold, and affects a different number of tokens from \method{}.

\subsection{Global Reweight}

Global Reweight is a batch-relative baseline. Let
\begin{equation}
    z_i = \mathrm{clip}\left(\frac{d_i-\mathrm{mean}(d)}{\mathrm{std}(d)}, -z_{\max}, z_{\max}\right).
\end{equation}
The token weight is
\begin{equation}
    w_i = \frac{\exp(-\alpha z_i)}{\mathrm{mean}_j\exp(-\alpha z_j)},
\end{equation}
followed by clipping to $[w_{\min}, w_{\max}]$. The adjusted divergence is $d'_i=d_i\,\mathrm{stopgrad}(w_i)$. We use $\alpha=0.3$, $z_{\max}=3.0$, $w_{\min}=0.25$, and $w_{\max}=2.0$. This method is useful as a contrast because it reweights all tokens by batch-relative divergence, while \method{} leaves non-extreme tokens unchanged.

\subsection{\method{}}

\method{} sets a detached dynamic threshold
$C=\mathrm{stopgrad}(k\,\mathrm{mean}_i(d_i))$ for each batch. We use $k=3.0$ in the main experiment. Tokens with $d_i\le C$ are unchanged. Tokens with $d_i>C$ use
\begin{equation}
    d'_i = d_i \frac{C}{\mathrm{stopgrad}(d_i)}.
\end{equation}
The forward value is capped at $C$, but the token still receives a nonzero gradient scaled by $C/d_i$. Detaching the threshold prevents gradients from flowing through the batch mean and matches the implementation.

\FloatBarrier
\section{Reproducibility Checklist}

Within each model family, all matched GKD variants share the same
initialization, teacher pair, data split, and anchored training recipe.
Student GKD training is repeated over three random seeds, and final
single-turn evaluation is then run on the resulting checkpoints. The following
items are fixed within each matched comparison unless a loss or support
intervention explicitly changes them:
initialization checkpoint, train/validation data, teacher routing, teacher API
type, maximum sequence lengths, batch size, learning rate schedule, rollout
mixture and sampling, distillation temperature, SFT-anchor weight, teacher
top-$K$ logit truncation, and evaluation datasets. The loss modifier is the only intended
difference among vanilla GKD, Hard Clip, Global Reweight, and \method{};
the support strategy changes only the distilled support and teacher probes.

For each run, we store final outputs, benchmark scores, training logs,
diagnostic samples, and the plotting inputs used to generate the paper figures.
The public artifact includes the modified GKD backend, sanitized
training-script templates, evaluation scripts, metric definitions, aggregate
results, and plotting scripts.

\FloatBarrier
\section{BFCL Multi-turn Loop Diagnostic}
\label{app:bfcl-multiturn}

The BFCL multi-turn diagnostic evaluates whether a model with a stronger tool-call prior becomes less usable in an interactive tool environment. It uses 800 BFCL v4 multi-turn tasks and 3,136 user turns sampled from the local processed BFCL multi-turn data. When the model emits a tool call, the harness returns a fixed simulated tool observation generated by the local BFCL-style harness and continues the dialogue until the model gives a non-tool final answer or reaches the maximum step limit. Invalid calls are counted when the model emits a tool-call marker that the harness cannot parse into a supported function call.

Tables~\ref{tab:appendix-multiturn-full} and
\ref{tab:appendix-multiturn-turn-correctness} report the full interaction and
correctness results before we define their components.

\begin{table}[H]
\centering
\scriptsize
\setlength{\tabcolsep}{2.5pt}
\begin{tabular}{lrrrrrr}
\toprule
Method & Calls/turn & Loop@3 & Loop@5 & Repeat call & Non-tool final & Dialogue exact \\
\midrule
Mixed SFT & 0.974 & 5.1 & 0.7 & 2.5 & 96.5 & 5.1 \\
Vanilla GKD & 1.515$\pm$0.122 & 15.1$\pm$3.0 & 8.8$\pm$3.5 & 17.7$\pm$6.4 & 87.3$\pm$7.2 & 7.0$\pm$0.3 \\
Hard Clip & 1.348 & 11.5 & 6.3 & 14.4 & 91.5 & 6.0 \\
Global Reweight & 1.297$\pm$0.088 & 10.8$\pm$1.4 & 5.5$\pm$1.3 & 13.1$\pm$2.7 & 93.6$\pm$1.6 & 5.3$\pm$0.9 \\
\method{} & 1.289$\pm$0.023 & 10.3$\pm$0.4 & 5.0$\pm$0.3 & 11.3$\pm$0.6 & 93.9$\pm$0.2 & 6.2$\pm$0.4 \\
\Supportunion{} & 1.118$\pm$0.057 & 8.0$\pm$0.9 & 3.3$\pm$0.5 & 8.9$\pm$1.0 & 92.3$\pm$0.6 & 4.1$\pm$0.3 \\
\bottomrule
\end{tabular}
\caption{Full BFCL multi-turn interaction results. Values except calls/turn
are percentages. ``Non-tool final'' records termination, not correctness.
Three-seed methods report mean$\pm$sample std; Hard Clip and Mixed SFT are
single runs.}
\label{tab:appendix-multiturn-full}
\end{table}

\begin{table}[H]
\centering
\scriptsize
\setlength{\tabcolsep}{2pt}
\begin{tabular}{lrrrrrr}
\toprule
Method & Required call & Supported req. exec/state & Empty-GT no-call & Turn prot. & Overall dlg. exact & Supported dlg. exact \\
\midrule
Mixed SFT & 79.77 & 29.46 & 54.85 & 32.56 & 5.12 & 6.83 \\
Vanilla GKD & 92.90$\pm$0.33 & 34.38$\pm$0.62 & 37.70$\pm$0.51 & 35.57$\pm$0.54 & 7.04$\pm$0.31 & 9.39$\pm$0.42 \\
Hard Clip & 88.88 & 32.69 & 44.90 & 34.55 & 6.00 & 8.00 \\
Global Reweight & 87.78$\pm$4.62 & 30.58$\pm$2.53 & 43.37$\pm$6.80 & 32.65$\pm$2.06 & 5.33$\pm$0.92 & 7.11$\pm$1.23 \\
\method{} & 88.50$\pm$2.01 & 32.15$\pm$0.80 & 42.96$\pm$6.13 & 33.93$\pm$1.17 & 6.17$\pm$0.38 & 8.22$\pm$0.51 \\
First-position support & 83.15$\pm$5.61 & 29.67$\pm$3.18 & 47.41$\pm$6.52 & 32.49$\pm$2.50 & 5.83$\pm$1.01 & 7.78$\pm$1.35 \\
All-position support & 73.24$\pm$4.62 & 24.76$\pm$1.80 & 61.49$\pm$4.54 & 28.60$\pm$1.50 & 3.38$\pm$0.50 & 4.50$\pm$0.67 \\
\bottomrule
\end{tabular}
\caption{Turn-checkpoint and dialogue-level correctness in the 9B BFCL
multi-turn diagnostic (percent). Required call and Empty-GT no-call use all
3,136 observed user turns. ``Supported'' metrics exclude the missing-function
category not fully implemented by the local harness. ``Turn prot.'' denotes
turn-level protocol success; ``dlg. exact'' denotes dialogue exact success.
Multi-seed rows report
mean$\pm$sample std; Mixed SFT and Hard Clip are single runs.}
\label{tab:appendix-multiturn-turn-correctness}
\end{table}

We report six interaction diagnostics plus overall dialogue exact success. \emph{Tool calls per turn} counts the average number of tool calls emitted per user turn. \emph{Loop@3} and \emph{Loop@5} measure whether the model keeps calling tools for at least three or five consecutive steps. \emph{Max-step hit} records turns that end by reaching the harness limit. Because this harness stops after five tool-call steps, Loop@5 and Max-step are numerically identical. \emph{Repeat same call} records repeated calls to the same function with the same arguments. \emph{Non-tool final answer} records whether the model eventually returns a non-tool response; it does not measure whether that response is task-correct.

For \emph{overall dialogue exact success}, we reuse the BFCL multi-turn execution/state
checker stored with each trajectory. A task first passes the required turns:
the checker executes model and reference calls from the same initial state and
requires the resulting observations and environment state to match. We then
apply the complementary irrelevance condition: every turn with empty ground
truth must contain no executable tool call. A task succeeds only if both
conditions hold for the complete dialogue. This is a custom-harness exact
metric, not an official BFCL leaderboard score, and it does not judge the
quality of a final natural-language response.

We additionally replay every saved executable call through the same checker
and score each observed user turn. \emph{Required-turn execution/state}
requires both the environment state and accumulated execution responses to
match at that turn's checkpoint. \emph{Turn-level protocol success} applies this
test on required turns and requires no executable call on empty-ground-truth
turns. The local harness did not issue BFCL's empty-message
function-supplementation prompt in the missing-function category. Its 200
placeholder positions are therefore excluded from observed-turn denominators,
and endpoint summaries marked \emph{supported} exclude that category. Offline
replay reproduces all 13,600 stored task-run checker decisions exactly.

Excluding the unsupported missing-function category raises vanilla dialogue
exact success from 7.04\% to 9.39\%, with 35.57\% turn-level protocol success;
fewer loops still do not improve endpoint correctness.